\documentclass{article}

     \usepackage[final,nonatbib]{neurips_2020}

\usepackage[utf8]{inputenc} 
\usepackage[T1]{fontenc}    
\usepackage{hyperref}       
\usepackage{url}            
\usepackage{booktabs}       
\usepackage{amsfonts}       
\usepackage{nicefrac}       
\usepackage{microtype}      

\usepackage{algorithm}
\usepackage{algorithmicx}
\usepackage[noend]{algpseudocode}

\usepackage{amsmath,amsthm}
\usepackage{enumitem}
\usepackage{xcolor}
\usepackage{graphicx}

\usepackage{bm}

\usepackage{tikz}
\usetikzlibrary{arrows}
\usetikzlibrary{arrows.meta}
\usetikzlibrary{shapes.misc, positioning}

\usepackage{caption}
\usepackage{subcaption}

\newcommand{\jussi}[1]{{{\color{blue} [JV: #1]}}}

\newcommand{\added}[1]{{{\color{black} #1}}}

\newtheorem{theorem}{Theorem}
\newtheorem{lemma}[theorem]{Lemma}
\newtheorem{proposition}[theorem]{Proposition}

\newcommand{\gadget}{\emph{Gadget}}
\newcommand{\baies}{\emph{Beeps}}
\newcommand{\bidag}{\emph{BiDAG}}

\newcommand{\be}{\begin{eqnarray}}
\newcommand{\ee}{\end{eqnarray}}
\newcommand{\bes}{\begin{eqnarray*}}
\newcommand{\ees}{\end{eqnarray*}}

\newcommand{\setminusx}{\!\setminus\!}

\newcommand{\parm}{\mathrm{pa}}
\newcommand{\pa}{\mathit{pa}}

\newcommand{\vect}[1]{\bm{#1}}
\newcommand{\matr}[1]{#1}
\newcommand{\tran}{^{\mkern-1.5mu\mathsf{T}}}

\newcommand{\precisionmat}{Q}
\newcommand{\precision}{q}

\newcommand{\C}{C}
\newcommand{\R}{R}
\newcommand{\x}{\vect{x}}
\newcommand{\e}{\vect{e}}
\renewcommand{\b}{\vect{b}}
\newcommand{\muu}{\boldsymbol{\mu}}
\long\def\comment#1{}

\title{
Towards Scalable Bayesian Learning of Causal DAGs
}
\author{%
  Jussi Viinikka\\
Department of Computer Science\\
University of Helsinki\\
\texttt{jussi.viinikka@helsinki.fi}
\And
Antti Hyttinen\\
HIIT \& Departiment of Computer Science\\
University of Helsinki\\
\texttt{antti.hyttinen@helsinki.fi}
\And  
  Johan Pensar\\
Department of Mathematics\\
University of Oslo\\
\texttt{johanpen@math.uio.no}
\And
Mikko Koivisto\\
Department of Computer Science\\
University of Helsinki\\
\texttt{mikko.koivisto@helsinki.fi}
}

\begin{document}

\maketitle

\begin{abstract}
We give methods for Bayesian inference of directed acyclic graphs, DAGs, and the induced causal effects from passively observed complete data. Our methods build on a recent Markov chain Monte Carlo scheme for learning Bayesian networks, which enables efficient approximate sampling from the graph posterior, provided that each node is assigned a small number $K$ of candidate parents. We present algorithmic \added{techniques} to significantly reduce the space and time requirements, which make the use of substantially larger values of $K$ feasible. Furthermore, we investigate the problem of selecting the candidate parents per node so as to maximize the covered posterior mass. Finally, we combine our sampling method with a novel Bayesian approach for estimating causal effects in linear Gaussian DAG models. Numerical experiments demonstrate the performance of our methods in detecting ancestor--descendant relations, and in causal effect estimation our Bayesian method is shown to outperform previous approaches. 
\end{abstract}

\section{Introduction}


Bayesian learning of graphical models aims at assigning any event of interest a posterior probability given observed data over the variables. In causal directed acylic graph (DAG) models, examples of such events include presence of a causal path between two variables and the total causal effect of one variable on another. While the posterior of the former event is quantified by a single number, the latter is represented by a distribution. \added{The Bayesian approach is particularly attractive in the causal setting due to its ability to properly account for the often non-negligible uncertainty in the inferred causal structure. In comparison, non-Bayesian structure learning methods are more limited in this aspect, as they typically return a single DAG, or Markov equivalence class, without any associated measure of uncertainty.} In the case of linear Gaussian models, the prospects of the Bayesian approach have recently been demonstrated \cite{Pensar2020,Castelletti2020}, showing an improved estimation accuracy over the original non-Bayesian IDA method \cite{IDA} and some of its later variants. However, the power of Bayesian learning stems from model averaging which unfortunately has appeared to be computationally intractable in the combinatorial space of DAGs. Hence, the currently existing and provably accurate algorithms are feasible only with up to around 25 variables \cite{Koivisto2004,Tian2009,Talvitie2019,Pensar2020}, and algorithms with somewhat looser accuracy guarantees to several dozens of variables \cite{Liao2019}. 

There have been several attempts to scale up Bayesian learning of graphical models using Markov chain Monte Carlo (MCMC). The first methods simulated a Markov chain on the space of DAGs by applying edge operations (add, remove, and reverse edge), yielding a sample of DAGs approximately from the posterior \cite{Madigan1995, Grzegorczyk2008}. To improve the sampler's ability to escape from local optima, subsequent works collapsed the space of DAGs to linear and partial node orderings covering multiple DAGs \cite{Friedman2003, Niinimaki16}. 
While the smaller state space and smoother posterior landscape enhanced the reliability of the order-based samplers, they still suffered from two major drawbacks. First, the sampling distribution is \emph{biased}, favoring graphs that are compatible with a larger number of orderings. This is particularly problematic in the causal setting, since the bias forces one to assign a nonuniform prior over equivalent DAGs. Markov chains directly on equivalence classes suffer, again, from the large, combinatorial state space \cite{Madigan1996,Castelletti2018}. Second, each simulation step is \emph{computationally expensive}, since it requires summing over the local scores of all order-compatible parent sets for each node. This issue is emphasized in linear Gaussian models, where also larger parent sets are more probable a priori, as the number of parameters for a node grows only linearly with the number of parents. 

The two issues were partly resolved in two recent works \cite{Kuipers2017,Kuipers2020}. The sampling bias was avoided by sampling ordered node partitions instead of node orderings. The per-step computational cost, in turn, was dramatically reduced by restricting the parents to a small candidate set (a technique also proposed earlier \cite{Friedman2003}) and, importantly, precomputing all possible score sums and storing them in a lookup table. Inspired by this progress, we here make several contributions to further advance the machinery and its applicability to causal inference. Specifically, we address the following questions.  
\begin{enumerate}
\item[Q1] \emph{How many candidate parents can we afford?} The number of candidate parents per node, $K$, is a critical parameter. We wish to let $K$ be as large as possible to cover well the space of DAGs; unfortunately, the memory requirements and preprocessing time grow exponentially in $K$. We present several algorithmic ideas to reduce the space and time requirements, and thereby, to allow for a substantially larger $K$; see Table~\ref{table:algresults}. \added{Put otherwise, for fixed, practical values of $K$ and the number of nodes $n$, the savings are by 2--3 orders of magnitude compared to previous work.}

\item[Q2] \emph{How to select the candidate parents?} The method assumes that we can select a moderate number of candidate parents per node, say $K = 15$, such that the posterior mass of DAGs is concentrated on the restricted family of DAGs, even if the number of nodes $n$ is much larger than $K$. We study to what extent this assumption holds by (i) formulating the selection task as an optimization problem, (ii) giving an exact algorithm to solve the problem optimally for moderate $n$, and (iii) introducing and empirically comparing various scalable heuristic algorithms to find good solutions when $n$ is large.   
\end{enumerate}

In addition to the above contributions and building upon our sampling method, we introduce a novel Bayesian approach for estimating causal effects in linear Gaussian DAG models with unknown causal structure, a subject of recent intensive ongoing research \cite{IDA,Maathuis10,stekhoven,Taruttis15,Pensar2020,Castelletti2020}.   

\begin{enumerate}
\item[Q3] \emph{How to obtain the posterior of causal effects?} In a Bayesian linear DAG model, the posterior of a causal effect is obtained by integrating over the unknowns (structure and parameters). We propose a three-stage sampling-based method to approximate the posterior: (i) we sample a DAG using our proposed sampling method, (ii) we sample the model parameters conditional on the DAG, and (iii) we map the model parameters to their implied causal effects using a matrix inversion technique. Importantly, the key novelty in our estimator compared to the IDA approach is to make use of the complete DAG structure in the estimation procedure. \added{ Figure~\ref{fig:example} shows example posterior distributions obtained by this method.}
\end{enumerate}

Like previous works~\cite{IDA}, we assume the data to be complete in the sense that there are no hidden variables (faithfulness and causal sufficiency). The scalability of our methods allows us to present the first empirical comparison of the Bayesian approach to non-Bayesian methods in higher dimensions.

\begin{table}[t!]
\centering
\caption{Space and time requirements with $n$ nodes and $K$ candidate parents per node}
{ \small 
\begin{tabular}{lccc}
\toprule
Task & Space & Time & Previous work \cite{Kuipers2020} \\ 
\midrule
Pre-processing & $O(3^K + 2^K n)$ & $O(3^K n)$ & $O(3^K n)$ space, $O(3^K K^2 n)$ time \\
Simulation step & $O(2^K n)$ & $O(n)$ & $O(3^K n)$ space\\
Sampling $r$ DAGs & $O(3^K + K n r)$ & $O(3^K n + K n r)$ & $O(2^K n r)$ time\\
\bottomrule
\end{tabular}
}
\label{table:algresults}
\end{table}

\section{Preliminaries}
\label{se:prelim}

We shall use the following notational conventions. For a tuple $(t_1, t_2, \ldots, t_k)$ we may write shorter $t_1 t_2 \cdots t_k$ or $(t_i)$, or just $t$. If $S$ is a set, we write $t_S$ for the tuple $(t_i : i \in S)$. 

A \emph{directed acyclic graph (DAG)} $(V, E)$ consists of a node set $V$ and an edge set $E \subseteq V \times V$ that contains no directed cycles. If $ij \in E$, call $i$ a \emph{parent} of $j$ and, conversely, $j$ a \emph{child} of $i$. Denote the set of parents of $j$ by $\parm_G(j)$, or by $\pa(j)$ when understood as a variable through the referred DAG. If there is a directed path from $i$ to $j$, call $i$ an \emph{ancestor} of $j$ and, conversely, $j$ a \emph{descendant} of $i$. 

For a vector of random variables $\x = x_1 x_2 \cdots x_n$, a \emph{Bayesian network (BN)} is a pair $(G, f)$, where $G$ is a DAG on the index set $V = \{1, 2, \dots, n\}$ and $f$ is a joint distribution that factorizes along $G$ as $f(\x) = \prod_{i=1}^n f(x_i | x_{\pa(i)})$. Specific representations of the conditional distributions yield more concrete models \cite{Koller09}. Among the most popular models are \emph{discrete BNs}, in which the support of each variable is finite with fully parameterized conditional probabilities, and \emph{linear Gaussian DAGs} \cite{Wright34,gh}, in which the local distributions are Gaussians. The latter corresponds to a structural equation model $\x := \muu+ \matr{B} (\x - \muu) +\e$, with $\e \sim \mathcal{N}( \mathbf{0}, \matr{\precisionmat} )$. Here $\precisionmat$ is a diagonal matrix of the error term precisions \added{and $B = (b_{ij})$ a matrix of edge weights}.  
The joint distribution of $\x$ is then $\mathcal{N}( \muu, \matr{W}  )$, with the precision matrix $\matr{W}=(\matr{I}-\matr{B})\tran\matr{\precisionmat}(\matr{I}-\matr{B})$. 

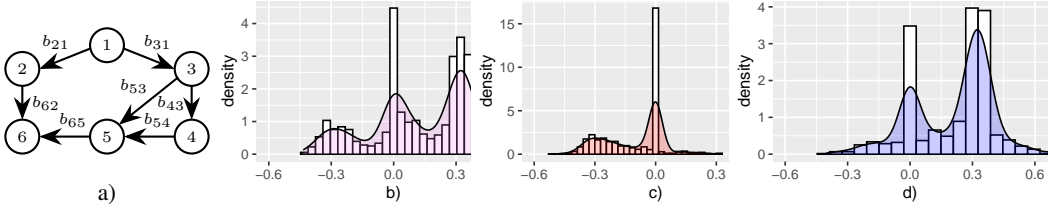
\begin{figure}[!t]
\centering
\tiny
\begin{tikzpicture}[scale=0.45]
\node at (5.0,-1.2) {{\footnotesize a)}};
\draw [black, thick] (2.5,0.5) circle (0.5);
\node at (2.5,0.5) {$6$};
\draw [black, thick] (5.0,0.5) circle (0.5);
\node at (5.0,0.5) {$5$};
\draw [black, thick] (7.5,0.5) circle (0.5);
\node at (7.5,0.5) {$4$};
\draw [black, thick] (7.5,2.5) circle (0.5);
\node at (7.5,2.5) {$3$};
\draw [black, thick] (5.0,3.2) circle (0.5);
\node at (5.0,3.2) {$1$};
\draw [black, thick] (2.5,2.5) circle (0.5);
\node at (2.5,2.5) {$2$};
\draw [-{Stealth[length=3mm]},thick] (5.5,3.1) to 
(7.0,2.5);
\node at (3.5,3.3) {$b_{21}$};
\node at (6.5,3.3) {$b_{31}$};

\node at (6.9,1.5) {$b_{43}$};
\node at (3.2,1.5) {$b_{62}$};

\node at (4.0,1.0) {$b_{65}$};
\node at (6.5,1.0) {$b_{54}$};
\node at (5.8,2.0) {$b_{53}$};
\node at (5.8,-0.95) {$\phantom{e}$};
\draw [-{Stealth[length=3mm]},thick] (4.5,3.1) to 
(3.0,2.5);
\draw [-{Stealth[length=3mm]},thick]  (4.5,0.5) to 
 (3.0,0.5) ;       
\draw [-{Stealth[length=3mm]},thick]  (2.5,2.0) to 
(2.5,1.0) ;                                                                                                                                                       
\draw [-{Stealth[length=3mm]},thick]  (7.0,0.5) to 
 (5.5,0.5) ;
\draw [-{Stealth[length=3mm]},thick] (7.5,2.0) to 
(7.5,1.0);
\draw [-{Stealth[length=3mm]},thick](7.1,2.25) to 
(5.35,0.85);
\end{tikzpicture} 
\hspace*{0pt}\includegraphics[width=0.31\textwidth]{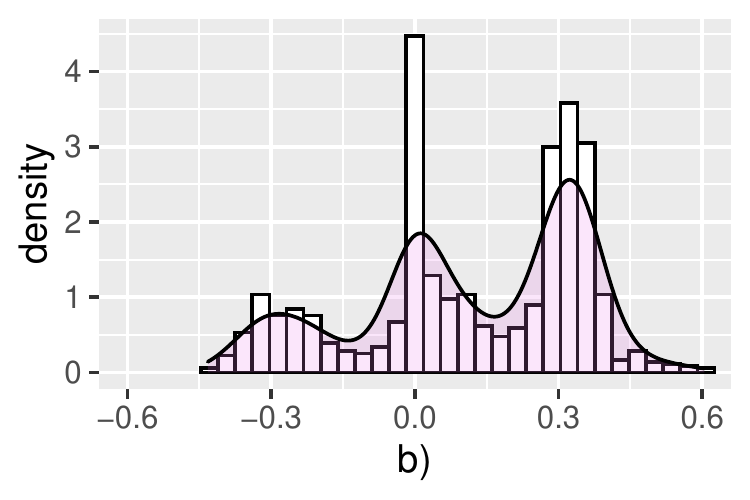} 
\hspace*{-27pt}\includegraphics[width=0.31\textwidth]{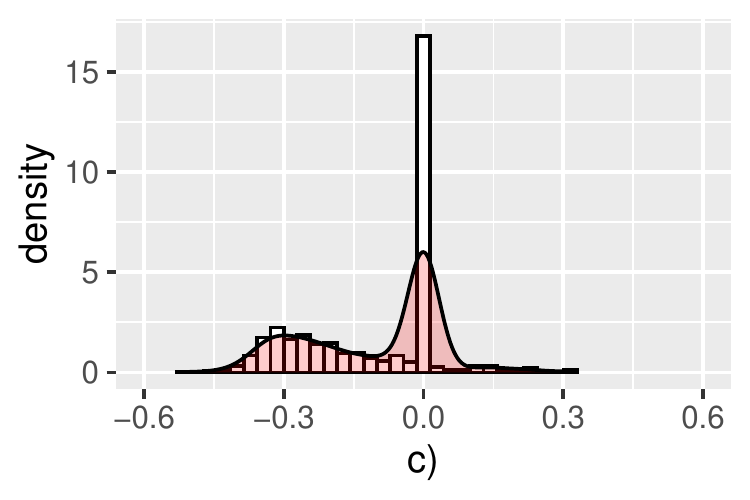} 
\hspace*{-27pt}\includegraphics[width=0.31\textwidth]{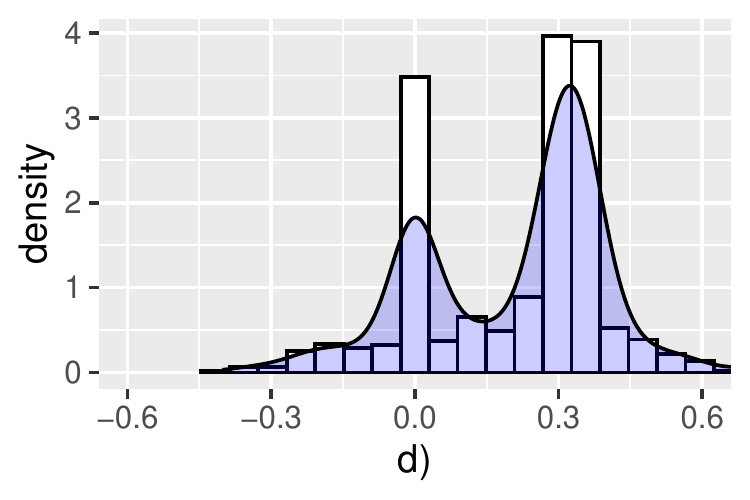} 

\caption{(a) A linear DAG model (error variances not shown).  (b--d) The posteriors of the linear causal effect of $x_1$ on $x_6$ given observational data, when intervening on $\{x_1\}$ in (b), $\{x_1,x_2\}$ in (c), and $\{x_1,x_3\}$ in (d). The posterior in (b) is a mixture of the  posteriors in (c) and (d).}
\label{fig:example}
\end{figure}

When a BN $(G, f)$ is interpreted as a causal model, $G$ encodes a hypothesis of the direction of causal relations. From $G$ alone, we can read off whether a node $j$ is an ancestor of $i$, and thus $x_j$ potentially has a causal effect on $x_i$. The magnitude is specified by the distribution $f$. We will focus on linear Gaussian DAGs, in which the \emph{causal effect} of $x_j$ on $x_i$ is quantified by a single scalar $a_{ij}$
obtained by summing up the weights of all directed paths from $j$ to $i$, the weight of a path equalling the product of the coefficients associated with the edges. In Figure~\ref{fig:example}(a), \added{node $1$ is an ancestor of node $6$} and $a_{61}=b_{62}b_{21}+b_{65}(b_{53}+b_{54}b_{43})b_{31}$.

To learn a BN $(G, f)$, we assume $N$ independent samples $\x_1, \x_2, \ldots, \x_N$ from $f$. We denote by $\matr{X}$ the $N \times n$ data matrix. We take a Bayesian approach and specify a joint distribution $p(G, f, \matr{X})$ as the product of the priors $p(G)$ and $p(f|G)$ and the likelihood $p(\matr{X} | G, f) = \prod_s f(\x_s)$. We assume the priors satisfy standard modularity properties, so that the posterior of $G$ can be written as 
\be \label{eq:Gpost}
	p(G|\matr{X}) \propto \pi(G) := \prod_{i=1}^n \pi_i\big(\parm_G(i)\big)\,,
	\quad \textrm{with} \quad
	\pi_i(S) := \rho_i(S)\, \ell_i(S)\,,
\ee
where $\rho_i$ and $\ell_i$ are factors of the DAG prior and the marginal likelihood: $p(G) \propto \prod_i \rho_i\big(\pa(i)\big)$ and $p(\matr{X} | G) = \prod_i \ell_i\big(\pa(i)\big)$. For example, in our experiments we put $\rho_i(S) = 1\big/\binom{n-1}{|S|}$ and composed the prior $p(f|G)$ from conjugate priors so that $\ell_i(S)$ admits a closed-form expression that is efficiently evaluated for any given node set $S$, and that yield the marginal likelihoods known as the \emph{BDe} and \emph{BGe scores} for discrete and Gaussian models, respectively. With these choices the posterior $p(G | X)$ is \emph{score equivalent}, meaning that the posterior probability is the same for Markov equivalent DAGs.  

\section{Scalable sampling of directed acyclic graphs} 
\label{se:sampling}

To draw a sample of DAGs approximately from the posterior distribution, we adopt the approach of Kuipers et al.\ \cite{Kuipers2017,Kuipers2020}, implemented in the \emph{BiDAG} package, with some major modifications. 

\subsection{Outline}

The basic idea is to sample DAGs by simulating a Markov chain whose stationary distribution is the posterior distribution. However, to enhance the mixing of the chain, we build a Markov chain on the smaller space of ordered partitions of the node set, each state being associated with multiple DAGs. 

Let $\R = R_1 R_2 \cdots R_k$ be an ordered set partition of $V$. We call $\R$ the \emph{root-partition} of a DAG $G$ if $R_1$ consists of the root nodes of $G$, $R_2$ consists of the root nodes of the residual graph $G - R_1$, and so forth; here $G - R_1$ is the graph obtained by removing from $G$ the nodes in $R_1$ and all incident arcs. Note that a DAG has a unique root-partition, whereas there may be several topological orders. For example, the root partition of the example DAG in Figure~\ref{fig:example}(a) is $\{1\} \{2,3\} \{4\} \{5\} \{6\}$.

The root-partition of $G$ is $\R$ exactly when every node in $R_1$ has zero parents and every node in $R_t$, with $t \geq 2$, has at least one parent from the previous part $R_{t-1}$ and the rest from the union $R_{1, t-1} := R_1 \cup R_2 \cup \cdots \cup R_{t-1}$. This is also evident in Figure~\ref{fig:example}(a). Thus, by the factorization \eqref{eq:Gpost}, the posterior probability of $\R$, i.e., the total probability of DAGs with root-partition $\R$, is given by  
\bes
	\pi(\R) = \prod_{t=1}^k \prod_{i \in R_t} \tau_i(R_{1,t-1}, R_{t-1})\,, \quad \textrm{with} \quad
	\tau_i(U, T) := \sum_{S \subseteq U : S \cap T \neq \emptyset} \pi_i(S)\,.
\ees
In words, $\tau_i(U, T)$ is the sum of local scores of node $i$ over all parents sets that contain at least one parent from $T$ and the rest from $U$. The factorization enables fast evaluation of $\pi(\R$), provided that the score sums $\tau_i(R_{1,t-1}, R_{t-1})$ can be computed fast. A fast evaluation is crucial for the scalability of the method, as the evaluation is required in every simulation step of the Markov chain. 

The key observation is the following \cite{Kuipers2020}. If node $i$ can only take parents from a small candidate parent set $C_i$, then it is feasible to precompute the needed values $\tau_i(U, T)$, for they only depend on the intersections $U \cap C_i$ and $T \cap C_i$. The evaluation then corresponds to a (nearly) constant-time table lookup. 
In Figure~\ref{fig:example}(a), we might discover that 
$C_{1}=\{2,3\}$,
$C_{2}=\{1\}$,
$C_{3}=\{1,4,5\}$,
$C_{4}=\{3,5\}$,
$C_{5}=\{3,4\}$, and
$C_{6}=\{2,5\}$ are good choices for the candidate parents by simple linear regression.
 
Finally, we generate DAGs conditionally on the sampled partitions. Generating a single DAG by enumerating all possible parent sets would require time $O(2^K n)$ \cite{Kuipers2020}, which is expensive. Instead, we will generate DAGs as postprocessing in time $O(K n)$ per DAG, by investing $O(3^K)$ space.

Algorithm~\ref{alg:dags} outlines the three phases of our method, we dub \gadget{} (Generating Acyclic DiGraphs Efficiently from Target). We describe the phases in more detail the remainder of this section.\footnote{
For the sake of exposition, we here consider simplifications of \emph{BiDAG} and \gadget{} that require all parents be from the $K$ candidates. In experiments we ran extended versions: \emph{BiDAG} additionally allows one parent outside the candidates, and \gadget{} any three or fewer parents; using known techniques \cite{Friedman2003,Niinimaki2013} this is still feasible.} 

\begin{algorithm}[!t]
\begin{algorithmic}[1]
\State \emph{Preprocessing.} 
Select a set of candidate parents $C_i$ for each node $i\in V$. Build a data structure that enables fast evaluation of $\tau_i(U, T)$ for any $i \in V, T \subseteq U \subseteq V\setminusx\{i\}$. 
\State \emph{Markov chain simulation.}
Generate a realization of a Markov chain $\R^0, \R^1, \ldots, \R^L$ whose stationary distribution is the posterior of root-partitions on $V$ using the Metropolis--Hastings algorithm. Store every $n$th sample $\R^s$.  
\State \emph{Postprocessing.}
Generate a DAG $G^s$ per sampled and stored $\R^s$.
\end{algorithmic} 
\caption{The \gadget{} method for sampling DAGs
\label{alg:dags}}
\end{algorithm}

\subsection{Preprocessing}

In what follows, we assume that each node $i$ is assigned a set of candidate parents $C_i$ of size $K$. We will consider the task of selecting the candidate parents for each node in Section~\ref{se:selection}. 

We aim at building a data structure that enables fast evaluation of the node-wise score sum $\tau_i(U, T)$ for any given $i, U, T$. To this end, for any $i \in V$ and $J \subseteq V\setminusx\{i\}$, let
\bes
	\tau_i(J) := \sum_{S \subseteq J \cap C_i} \pi_i(S)\,,
\ees
the sum of all local scores for node $i$ with parents from $J \cap C_i$. Clearly, $\tau_i(J) = \tau_i(J \cap C_i)$. Furthermore, the values $\tau_i(J)$ are sufficient for instant evaluation of a score sum, by subtraction: 
\begin{lemma}
Let $i \in V$ and $T \subset U \subseteq V\setminusx\{i\}$. Then 
$\tau_i(U, T) = \tau_i(U) - \tau_i(U\setminusx T)$.
\end{lemma}
(Indeed, if $S \subseteq U$, then either $S$ intersects $T$ or $S \subseteq U\setminusx T$, implying $\tau_i(U) = \tau_i(U, T) + \tau_i(U\setminus T)$.) 

Put together, it suffices to precompue for each node $i$ the values $\tau_i(J)$ for all $J \subseteq C_i$. Since $\tau_i$ is the zeta transform of $\pi_i$ over the subset lattice of $C_i$, it can be computed in time $O(2^K K)$; see Supplement~A.1. The space requirement is $O(2^K)$ per node. This improves upon a brute-force approach, which requires building time $O(3^K K^2)$ and storage size $O(3^K)$ per node \cite{Kuipers2020}.

When the arithmetic is with fixed-precision numbers, there is a risk of so-called catastrophic cancellation. That is, the outcome of a subtraction may vanish (due to limited precision), even if the exact value is non-zero. While such cases occured only rarely in our experiments, we build a secondary data structure to handle them; if there are $m$ cases, the construction takes $O(3^K n)$ time and $O(3^K + m)$ space (Suppl.~A.2). Note: in Table~\ref{table:algresults} we made the mild assumption that $m = O(2^K n)$. 

\subsection{Markov chain simulation}

We follow the partition MCMC method \cite{Kuipers2017,Kuipers2020} and simulate a Markov chain $\R^1, \R^2, \ldots, \R^L$ of some appropriate length $L$ on ordered set partitions of $V$ using the Metropolis--Hastings algorithm. At state $\R^s$ a candidate $\R'$ for the next state is generated by either splitting a part, merging two adjacent parts, or swapping nodes in different parts, uniformly at random over the valid choices; denote this distribution by $q(\R'|\R^s)$. The proposal is accepted as the new state $\R^{s+1}$ with probability $\min\{1, \alpha\}$, where $\alpha = \pi(\R')/\pi(\R^s) \times  q(\R^s|\R')/q(\R'|\R^s)$; otherwise $\R^{s+1}$ is set to $\R^s$.  

Instead of simulating a single long chain, we enhance the mixing of the chain by employing Metropolis coupling \cite{Geyer1991}: we run $M > 1$ shorter ``heated'' chains in parallel, the $k$th chain with stationary distribution proportional to $\pi^{k/M}$. In every other step, two chains $k$ and $l = k+1$ are selected uniformly at random, and a swap of their states, $\R^{s,k}$ and $\R^{s,l}$, is proposed; the acceptance ratio $\alpha$ equals the $M$th root of $\pi(\R^{s,k}) / \pi(\R^{s,l})$. In our experiments, we put $M := 16$. 

\subsection{Postprocessing}\label{se:postprocessing}

We generate a DAG per sampled partition as postprocessing, in order to save space. The key observation is that, instead of generating an entire DAG for each partition in turn, we can proceed one node in turn, and generate the parent sets of the node for all the DAGs we are generating. This ``transposition trick'' enables reusing the space we need for efficient sampling of parent sets. \added{Furthermore, for sampling the parent sets of a fixed node, we introduce a data structure to index certain weighted sums, enabling efficient sampling of constrained sets. }

Recall that the root-partition of the DAG in Figure~\ref{fig:example}(a) is $\{1\}\{2, 3\}\{4\}\{5\}\{6\}$. Now, consider generating a random DAG compatible with this partition. Since each node must take at least one parent from the predecessor part, we must include the edges $5\rightarrow 6$, $4\rightarrow 5$, $1\rightarrow 2$ and $1\rightarrow 3$. In addition, either $2\rightarrow 4$ or $3\rightarrow 4$ is included. The parent sets will be sampled according to the scores $\pi_i$ as explained below such that these restrictions are satisfied.

For a more technical description, consider generating a DAG $G$ from the posterior distribution given that the root-partition of $G$ is $\R$. We can draw $G$ by sampling independently for each node $i \in R_t$ a parent set $S \subseteq R_{1, t-1}$ that intersects $R_{t-1}$, with probability proportional to $\pi_i(S)$. If implemented in a direct way, this takes time $O(2^K)$ per node, but no additional space \cite{Kuipers2020}. 

We reduce the time requirement to $O(K)$, by investing $O(3^K)$ preparation time per node and $O(3^K)$ space in total. Consider a fixed node $i$. The idea is to construct a data structure that, given any node sets $T \subseteq U \subseteq C_i$, enables drawing a parent set $S \subseteq U$ that intersects $T$, with probability proportional to $\pi_i(S)$. We draw $S$ in $O(|U|)$ iterative steps, in each step deciding whether a particular node $j \in U$ is included in $S$ or not. To enable this, our data structure stores the sum of $\pi_i(S)$ over $T' \subseteq S \subseteq U'$ for all pairs $T' \subseteq U' \subseteq C_i$; see Supplement~A.3 for details.  

If the number of sampled DAGs is $r$, the total space and time requirements of postprocessing are $O(3^K + K n r)$ and $O(3^K n + K n r)$, respectively. In contrast to the brute-force approach \cite{Kuipers2020}, our trick makes it feasible to sample large numbers of DAGs.

\section{Selection of candidate parents}
\label{se:selection}

We wish to find a good set of $K$ candidate parents for each node. Our interest is in algorithms that scale up to hundreds of nodes. While we cannot expect an algorithm that always returns an optimal set, we can hope for a heuristic that finds sets covering a large fraction of the graph posterior mass. We formalize this problem, consider the issue of evaluating the performance of a given algorithm, and finally, briefly describe several alternative algorithms and report on an empirical study. 

\subsection{The maximum coverage problem}

Consider a tuple of candidate parent sets $\C = C_1 C_2 \cdots C_n$. Define the \emph{coverage} of $\C$ as the posterior probability that the parents of $i$ belong to $C_i$ for all nodes $i$. Likewise, define the \emph{mean coverage} of $\C$ as the average of the marginal posterior probabilities that the parents of $i$ belong to $C_i$. 

Given a $\C$, we can compute the coverage and mean coverage in time $O(3^n n)$ and space $O(2^n n)$. Namely, within this complexity we can evaluate the partition function \cite{Tian2009} as well as the marginal posterior probabilities of all the $2^{n-1}$ possible parent sets of each node \cite{Pensar2020}. Thus exact evaluation of a given $\C$ is computationally feasible up to around $n = 22$. 

The \emph{maximum (mean) coverage} problem is to find a $\C$ so as to maximize the (mean) coverage, subject to the constraint $|C_i| \leq K$ for all $i$. 
The mean variant is tractable for small $n$: 
\begin{proposition}\label{prop:opt}
The maximum mean coverage problem can be solved in time $O(3^n n)$.
\end{proposition}
\vspace*{-12pt}
\begin{proof}
Compute first the marginal posterior probabilities $g_i(S) := p(\pa(i)=S|X)$ for all $S \subseteq V\setminusx\{i\}$ in time $O(3^n n)$ \cite{Pensar2020}. 
Then compute $g'_i(T) := \sum_{S \subseteq T} g_i(S)$ for all $T \subseteq V\setminusx\{i\}$ in time $O(3^n n)$. Finally, for each $i$ return a $K$-set $C_i$ that maximizes $g'_i(C_i)$; this takes time $O(2^n n)$.     
\end{proof}

\subsection{Scalable algorithms for the maximum coverage problem}

\begin{table}[t!]
\centering
\caption{Algorithms for selecting the candidate parents of node $i$}
{ \small 
  \setlength{\tabcolsep}{4pt}
\begin{tabular}{rl}
\toprule
\emph{Opt} & Select a $K$-set $C_i$ so as to maximize the posterior probability that $\pa(i) \subseteq C_i$ (cf.\ Prop.~\ref{prop:opt})\\
\emph{Top} & Select the $K$ nodes $j$ with the highest local score $\pi_i(\{j\})$\\  
\emph{PCb} & Merge the neighborhoods of $i$, \added{excluding children}, returned by \emph{PC} on $20$ bootstrap samples\\
\emph{MBb} & Merge the Markov blankets of $i$ returned by \emph{IA} on $20$ bootstrap samples\\
\emph{GESb} & Merge the neighborhoods of $i$, \added{excluding children}, returned by \emph{GES} on $20$ bootstrap samples \\
\emph{Greedy} & Iteratively, add a best node to $C_i$, initially empty; the \emph{goodness} of $j$ is $\max_{S \subseteq C_i} \pi_i(S\cup\{j\})$\\
\emph{Back\&Forth} & Starting from a random $K$-set delete a worst and add a best node, alternatingly, until the same\\ 
\bottomrule
\end{tabular}
}
\label{table:selectionalgs}
\end{table}

\begin{figure}[b!]	
  \centering
\hspace*{-15pt}
  \begin{subfigure}[t]{.30\textwidth}
    \includegraphics[scale=0.475]{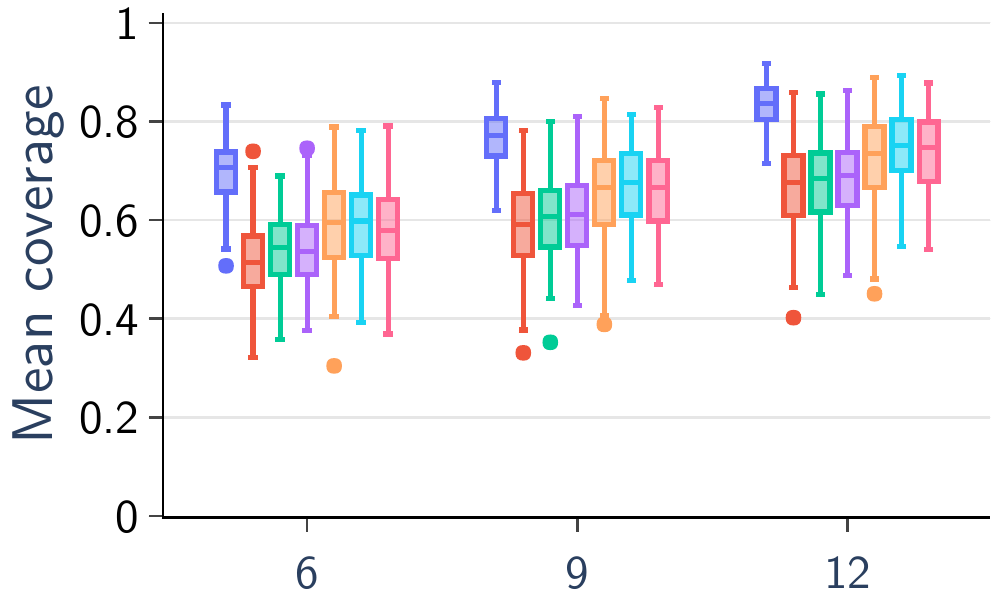}
    \caption{Gaussian, $N = 50$}\label{fig:coverage:G50}		
  \end{subfigure}
\hspace*{10pt}
  \begin{subfigure}[t]{.30\textwidth}
    \includegraphics[scale=0.475]{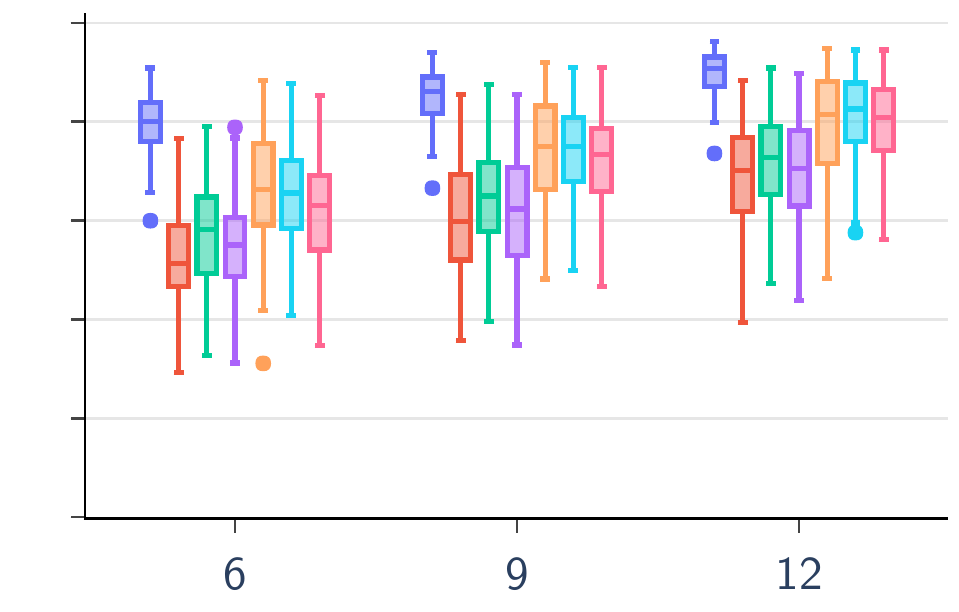}
    \caption{Gaussian, $N = 200$}\label{fig:coverage:G200}
  \end{subfigure}
\hspace*{0pt}
  \begin{subfigure}[t]{.30\textwidth}
    \includegraphics[scale=0.475]{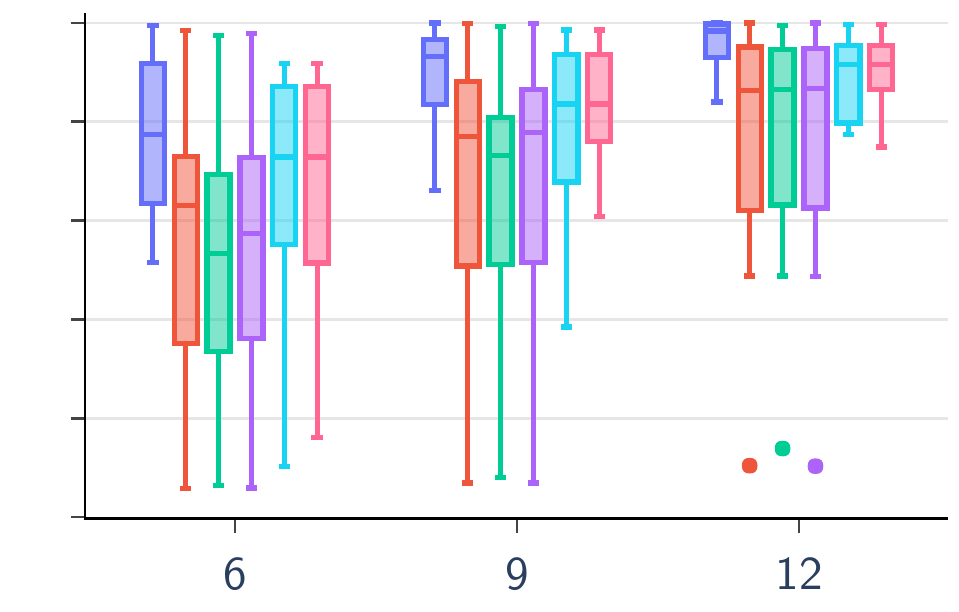}
    \caption{Discrete, varying $N$}\label{fig:coverage:discrete}
  \end{subfigure}
  \medskip

  \begin{subfigure}[t]{1.0\textwidth}
    \centering
    \includegraphics[scale=0.675]{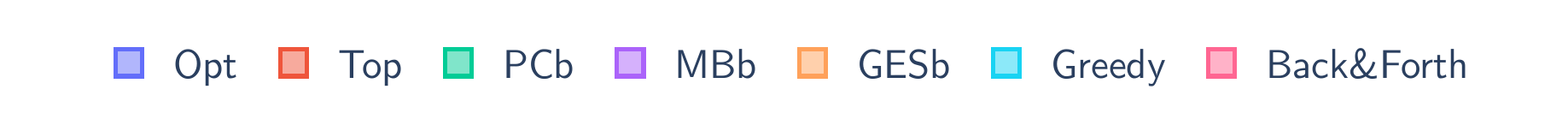}
  \end{subfigure}  

  \caption{Performance comparison on selecting  $K = 6, 9, 12$ candidate parents with (a, b) synthetic data over $20$ nodes and (c) benchmark data sets over $17$--$23$ nodes with $101 \leq N \leq 8124$ data points.  
}\label{fig:12e2e23jife}
\label{fig:selectionplots}
\end{figure}

For larger numbers of nodes $n$, we have to resort to faster algorithms that are only guaranteed to find a locally optimal collection of candidate parent sets. We tested several heuristics, summarized in Table~\ref{table:selectionalgs} (details in Suppl.~C). Some rely on existing sophisticated algorithms for finding the Markov equivalence class (the \emph{PC} algorithm, using independence tests \cite{Spirtes:1993, colombo:2014}; greedy equivalence search, \emph{GES} using the BIC score \cite{chickering:2002}) or the Markov blanket of a target node (the \emph{Incremental Association} algorithm, \emph{IA} \cite{tsamardinos:2003}) of the unknown DAG; we ran the basic algorithms on $20$ bootstrap samples of the data, took the union of the returned neighborhoods, and removed or added the lowest- or highest-scoring nodes to get exactly $K$ candidates. Other algorithms are more elementary and handle each node separately, considering parent sets that are either singletons or subsets of an already constructed candidate set. Our implementations build on standard software \cite{scutari:2010, kalisch:2012, hauser:2012, Bartlett2017}.
 
For an empirical comparison of the heuristics, we set $n$ to $20$ to enable exact evaluation of the achieved coverage and comparison to the best possible performance (\emph{Opt}, cf.~Prop.~\ref{prop:opt}). We sampled two data sets of size $N = 50$ and $N = 200$ from each of 100 synthetic linear Gaussian DAGs, generated so that the expected neighborhood size was $4$, the edge coefficients and the variances of the disturbances uniformly distributed on $\pm[0.1, 2]$ and $[0.5, 2]$, respectively. We observe that the coverage of optimal sets of $K$ candidates increases with $K$ and $N$, reaching $0.90$ on average at $K = 12$ and $N = 200$ (Fig.~\ref{fig:selectionplots}(a, b)). \emph{Greedy} is the best of the heuristics and gets the closer to \emph{Opt}, the larger the size $K$.  

To investigate the performance on discrete real data, we also ran the algorithms on 8 data sets obtained from the UCI machine learning repository \cite{Dua:2019}, with up to 23 variables, using available preprocessed sets \cite{Malone:2018}. \added{In Fig.~\ref{fig:selectionplots}(c)}, we observe that \emph{Greedy} and its \emph{Back\&Forth} variant achieve coverages close to \emph{Opt}; the other algorithms perform worse. \added{\emph{GESb} is not shown for discrete data, as the employed software only allowed Gaussian BIC to be used. }

\comment{\jussi{
  new section ``5. Experiments on Markov chain mixing''
  


- Same data as in "Experiments on causal inference"? Just for Gaussian then.

- Some words on what can be seen in the traces.
}}

\section{Bayesian estimation of linear causal effects}
\label{se:causal}

The ability to sample DAGs (approximately) from the posterior distribution offers us a way to sample (pairwise) causal effects from the posterior distribution in linear Gaussian models. Algorithm~\ref{alg:ce2} outlines our method, dubbed \baies~(Bayesian Effect Estimation by Posterior Sampling).


\begin{algorithm}[!t]
\begin{algorithmic}[1]
\State   Sample DAGs $\{G^s\}_{s=1}^L$ approx.\ from the posterior $p(G|X)$, e.g., using \gadget{} (Section~\ref{se:sampling}).
\State For each $G^s$, sample $B^s$ from the posterior $p(B | G^s, X)$, each row independently (Eq.~\ref{eq:t}).
\State For each $B^s$, compute the matrix of pairwise causal effects $A^s$ via $A=(I-B)^{-1}$.
\State Output $\{A^s\}_{s=1}^L$.
\end{algorithmic} 
\caption{The \baies~method for sampling from the posterior of linear causal effects 
\label{alg:ce2}}
\end{algorithm}

Our goal is to draw a sample from the posterior $p(A | X)$, where $A = (a_{ij})$ is the matrix of pairwise causal effects and $X$ the data. Conveniently, $A$ can be expressed as a converging geometric series w.r.t. the edge weight matrix $B$, resulting in $A = (I-B)^{-1}$. Using this relation, $A$ can readily be computed from samples of $B$ drawn from the posterior $p(B | X)$. To draw $B$, we view $p(B | X)$ as a marginal of $p(B, G | X)$, and by the chain rule, draw first $G$ from $p(G|X)$ and then $B$ from $p(B | G, X)$. In what follows, we assume that $G$ has already been sampled and focus on the latter task.

Recall that we parameterize our linear Gaussian DAG by the mean vector $\muu$, the matrix  $B$, and the diagonal matrix of error term precisions $\precisionmat$. Geiger and Heckerman~\cite{gh,Geiger02} showed that there is a unique class of priors over these parameters satisfying the desirable properties of global and local modularity and the score-equivalence of the marginal likelihood $p(X | G)$, the BGe score. Moreover, for any prior from this class, we obtain the posterior of $B$ analytically: the rows of $B$ are independent with a $t$-distribution whose parameters can be efficiently computed. Since some of the key formulas in the literature contain small errors and typos; we give a complete derivation below and in Supplement~B. 

We begin with a normal--Wishart prior on the parameterization by $\muu$ and the precision matrix $W$: 
\bes
	\muu \,|\, W \sim \mathcal{N}( \vect{\nu}, \alpha_\mu W )\,, 
	\qquad  W\sim  \mathcal{W} (   T^{-1}, \alpha_w  )\,.
\ees
Here the scalars $\alpha_\mu$, $\alpha_w$, vector $\vect{\nu}$, and matrix $T$ are hyperparameters, which do not depend on the DAG $G$.\footnote{With the notation $\alpha_\mu$ and $\alpha_w$ we adhere to the choices in the key references \cite{Geiger02,kuipers2014}.} By change of variables, this is transformed to a prior over $\muu$, $B$ and $\precisionmat$, conditional on $G$ \cite{Geiger02}. After integrating out $\muu$, the marginal prior $p(B,\precisionmat | G)$ factorizes, due to global and local parameter modularity, into a product of $p(\b_i,\precision_i | \pa(i))$ over the nodes $i$; here  $\b_i$ is the $i$th row of $B$ and $\precision_i $ the $i$th diagonal element of $\precisionmat$. The prior for $\b_i$ and $\precision_i$ given $pa(i)$ is then (see Suppl.~B)  
\bes
	\b_i \,|\, \precision_i\,  \sim \mathcal{N}\big(\, (T_{11})^{-1}T_{12}\,,\; \precision_iT_{11}\,\big)\,,
	\qquad 
	\precision_i \sim \mathcal{W} \big( \, (T_{22}- T_{21}(T_{11})^{-1}T_{12} )^{-1}\,,\; \alpha_w-n+l\, \big)\,,
\ees
where the blocks of $T$ are $T_{11}:=T[\pa(i),\pa(i)]$, $T_{12}:=(T_{21})\tran = T[\pa(i),i]$, $T_{22}:=T[i, i]$, and $l-1$ is the number of parents of $i$. This corrects some errors in the formulas of Geiger and Heckerman \cite{Geiger02} for the degrees of freedom (noted also by Kuipers et al.\ \cite{kuipers2014}) and some typos in the matrices. 

The posterior is of exactly the same form, just  $\alpha_w$ and $T$ replaced, respectively, by 
\bes
	\alpha_w': = \alpha_w+N
	\quad \textrm{and} \qquad 
	R := T +S_N+\frac{\alpha_\mu N}{\alpha_\mu+N}(\vect{\nu} - \vect{\bar{x}}_N)(\vect{\nu} - \vect{\bar{x}}_N)\tran\,, 
\ees
where $\vect{\bar{x}}_N := \frac{1}{N}\sum_{s}\x_s$ and $S_N=\sum_{s} (\x_s-\vect{\bar{x}}_N)(\x_s-\vect{\bar{x}}_N)\tran$. 
Finally, integrating out $\precision_i$ yields 
\be
	\b_i \,|\, X, \pa(i) \,\sim\, t\Big( 
	(R_{11})^{-1}R _{12}\,,\;   
	\frac{\alpha_w'-n+l}{ R_{22}- R_{21}(R_{11})^{-1}R_{12} }\, R_{11}\,,\;
	\alpha_w'-n+l\Big)\,. \label{eq:t}
\ee

\medskip
\baies~differs from IDA-based methods (including the Bayesian ones,  \emph{BIDA}~\cite{Pensar2020} and \emph{OBMA}~\cite{Castelletti2020}), which estimate the causal effect from $x$ to $y$ by a linear regression of $y$ on $x$ and the parents of $x$. In contrast,  \baies~takes into account the whole graph structure and estimates of the single coefficients. This improves accuracy: e.g., the estimate is \emph{always exactly} zero when $x$ is not an ancestor of $y$. Furthermore, our method enables estimation of causal effects under multiple interventions~\cite{joint} by replacing the coefficients into the intervened variables in $B$ with zero in Step~3 of Algorithm~\ref{alg:ce2}.

\section{Experiments on causal inference}
\label{se:experiments}

%

\begin{figure}[t!]
\centering
  \hspace*{-16pt}  
  \begin{subfigure}[t]{.26\textwidth}
\includegraphics[height=100pt,trim={0 0 30pt 0},clip]{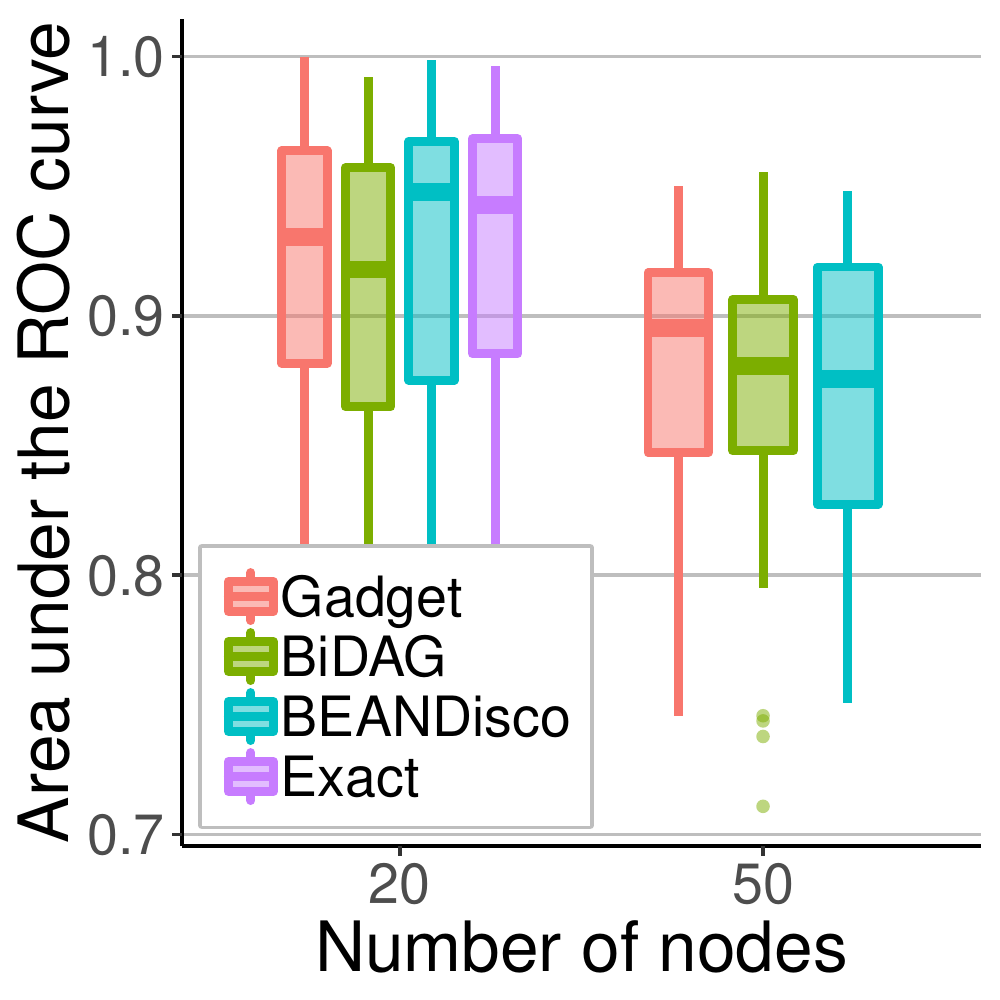}
    \caption{Ancestor relations}\label{fig:auroc}		
  \end{subfigure}
  \hspace*{-16pt}
  \begin{subfigure}[t]{.26\textwidth}
\includegraphics[height=100pt,trim={0 0 55pt 0},clip]{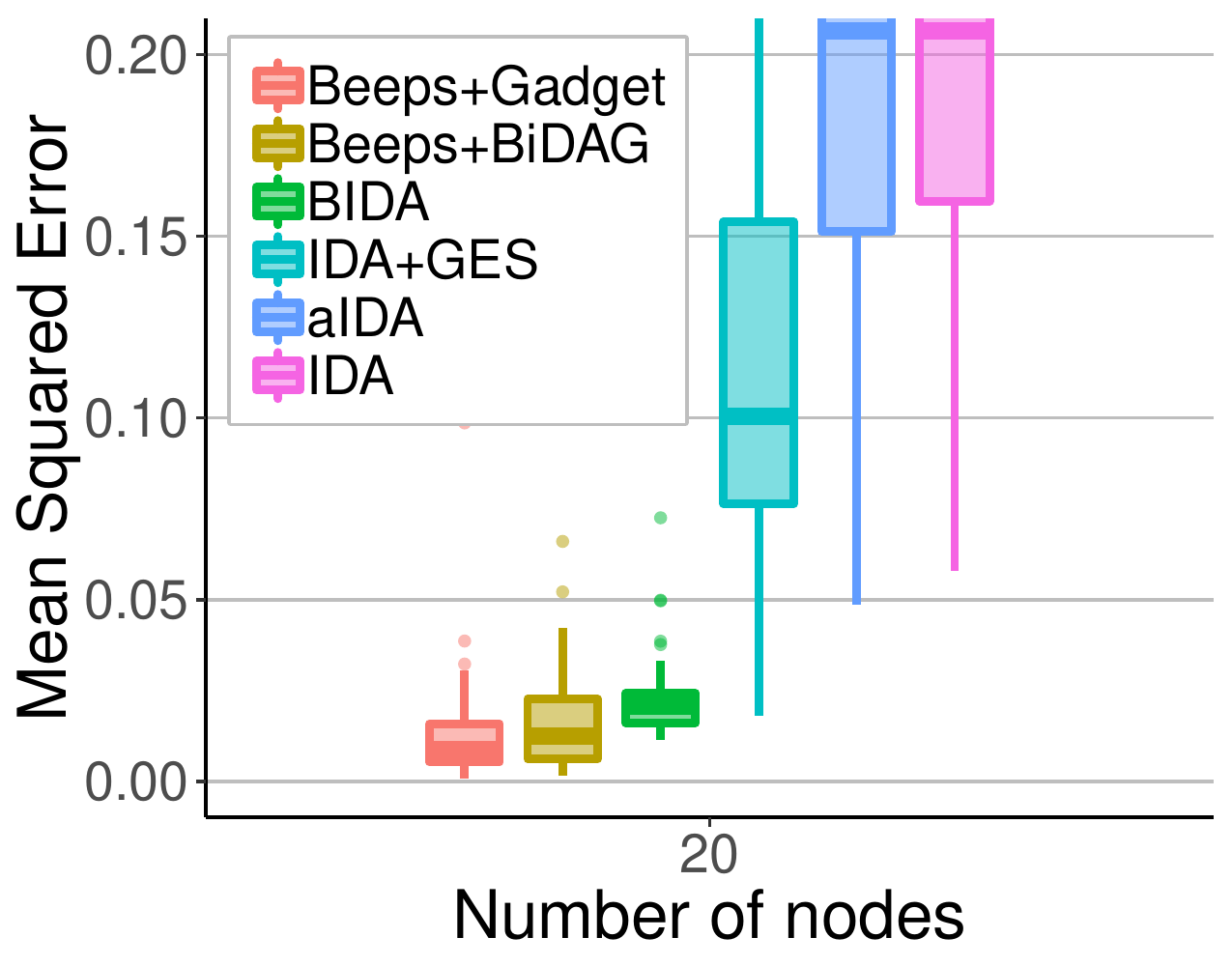}
    \caption{Causal effects}\label{fig:bida}
  \end{subfigure}
\hspace*{-5pt}
  \begin{subfigure}[t]{.26\textwidth}
\includegraphics[height=100pt,trim={0 0 55pt 0},clip]{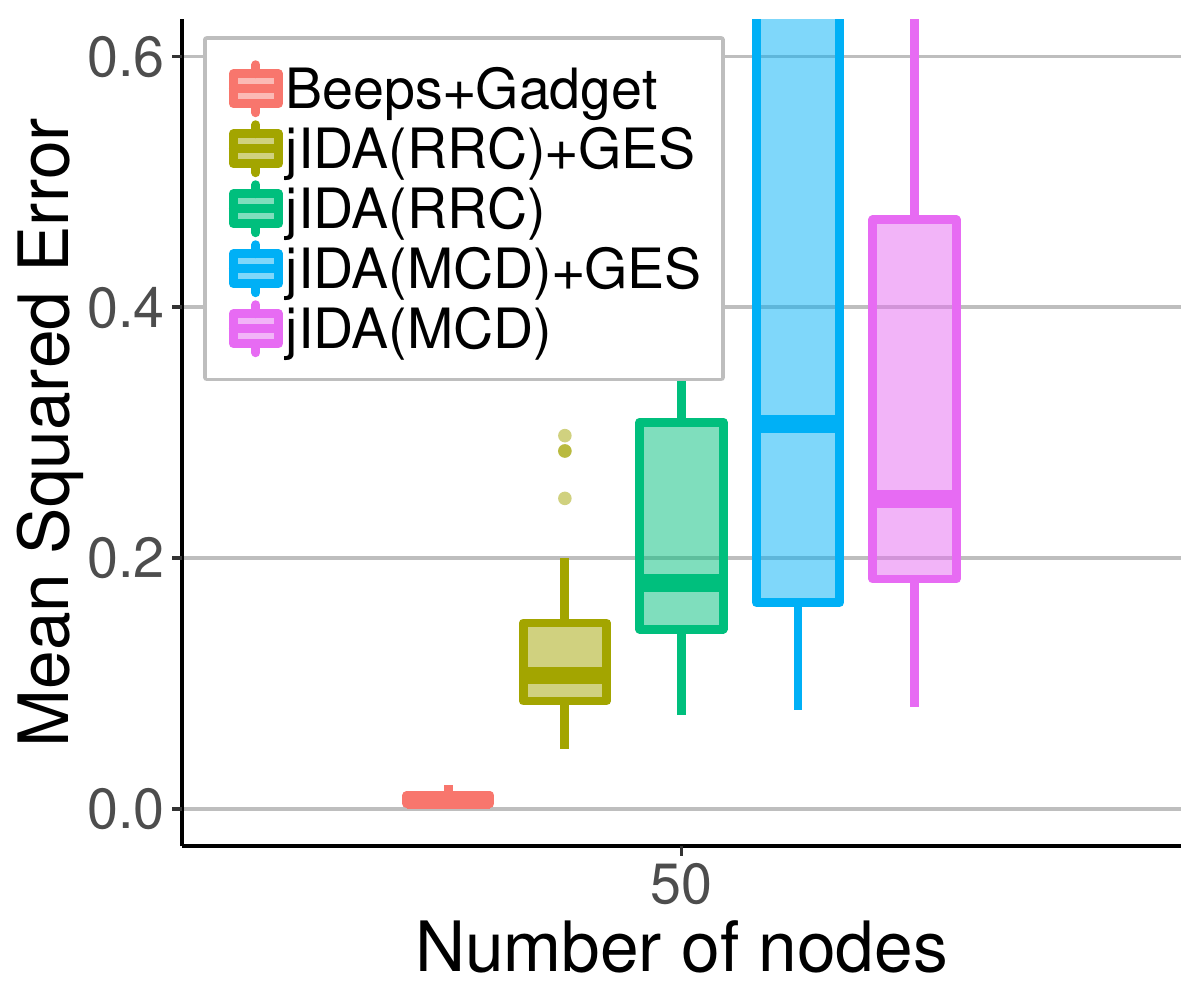}
    \caption{Joint causal effects}\label{fig:joint}
  \end{subfigure}    
\hspace*{-10pt}
  \begin{subfigure}[t]{.22\textwidth}
\includegraphics[height=100pt,trim={0 0 0 0},clip]{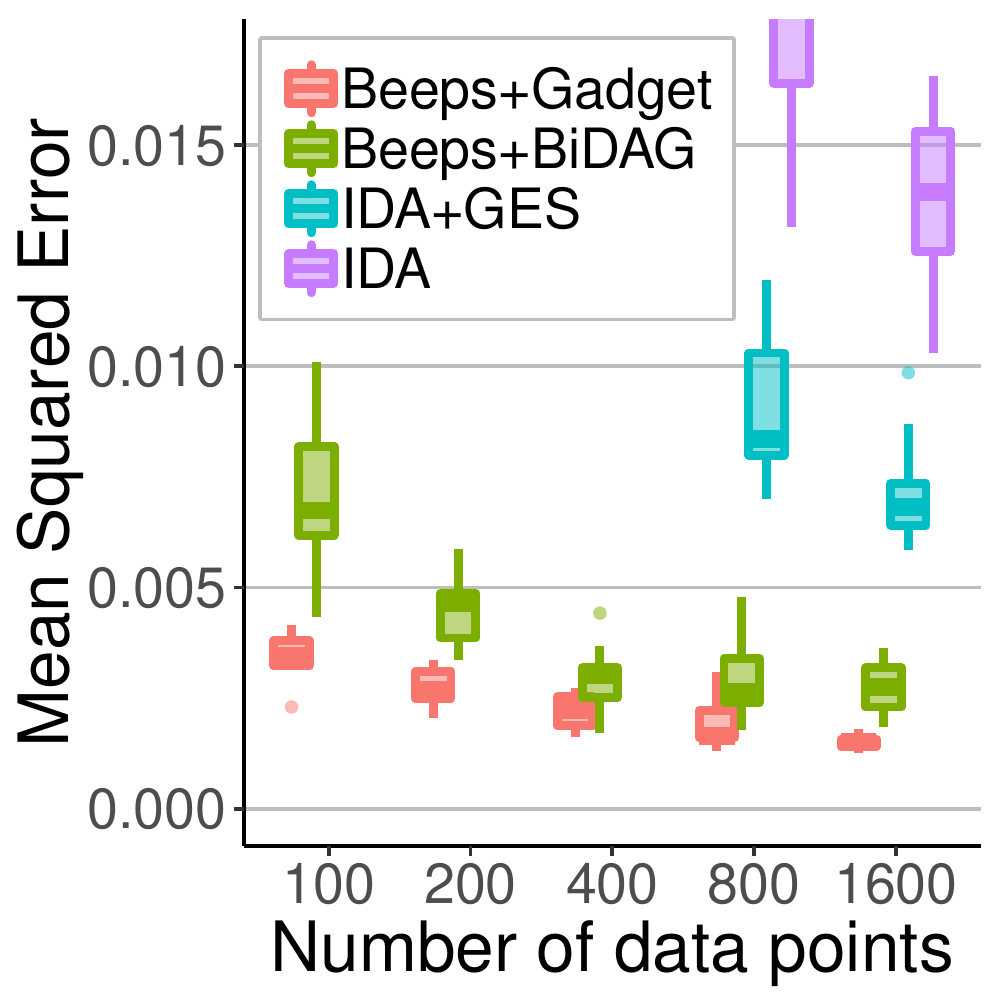} 
    \caption{107-node ARTH150}\label{fig:ara} 
  \end{subfigure}    

\caption{Performance comparisons. 
(a) Bayesian methods on inferring ancestor relations from 
 discrete data. 
Estimating (b) marginal and (c) joint causal effects from 
Gaussian data.      
(d) Estimating causal effects from data sets from a benchmark BN. The MCMC methods were ran for $1$ and $3$ hours for the $20$- and $50$-node data, respectively, and $12$ hours for (d); the other methods were faster. 
}\label{fig:mainplots}
\end{figure}

We compared our algorithms\footnote{\added{We provide a Python interface for both algorithms, with many time critical parts implemented in C++. For source code see https://www.cs.helsinki.fi/group/sop/gadget-beeps.}} to state-of-the-art Bayesian and non-Bayesian methods for discovering ancestor relations and estimating causal effects (marginal and joint). For a complete set of results and the choices of the various user parameters of the methods, we refer to Supplement~D.

We first evaluated the efficiency of \gadget{} in sampling DAGs from the posterior and detecting ancestor relations. We considered data on $20$ and $50$ nodes to enable comparison to an exact algorithm (for $n = 20$) \cite{Pensar2020} and two state-of-the-art MCMC methods, \emph{BEANDisco} \cite{Niinimaki16} and \emph{BiDAG} \cite{Kuipers2020}. We generated $400$ data points from 50 binary BNs with av.\ neighbourhood size 4. We observe that  \gadget{} closely matches or outperforms the other MCMC methods and the exact algorithm \added{(Fig.~\ref{fig:mainplots}(a))}. 

We then evaluated the performance of our \baies~method in estimating causal effects, using either \gadget{} or \emph{BiDAG} as the DAG sampler. To enable an informative comparison to the state-of-the-art scalable methods, i.e., variants of the IDA method~\cite{IDA}, we condense the effect estimates  to the mean value and calculate the mean-squared error~\cite{Pensar2020,Castelletti2020}. We generated 200 data points from 50 Gaussian BNs with neighourhood size 4. Our method achieves better accuracy in causal effect estimation compared to the BIDA method, which uses exact computation (but a different effect estimation technique (Fig.~\ref{fig:mainplots}(b)). We evaluated the performance of \baies~also in estimating joint causal effects (Fig.~\ref{fig:mainplots}(c)).  Our method clearly outperforms the available IDA-based methods~\cite{joint} in accuracy.

Finally, we obtained $50$ datasets with $100$--$1600$ data points from a benchmark Gaussian BN on gene expressions of Arabidopsis thaliana with $n = 107$ nodes~\cite{smith2004diurnal,DBLP:journals/bmcsb/Opgen-RheinS07}. We ran the MCMC  methods  12 hours or up to $10^8$ MCMC iterations. Despite data from a single source, the performance of \emph{BiDAG} varies considerably: for 200-400 data points it can often reach the limiting $10^8$ iterations but for 800 and 1600 data points \emph{BiDAG} fails to complete 100 iterations for $4/10$ and $8/10$ datasets respectively. \emph{Gadget} is able to use $K=15$ candidate parents for all data sets, and with \baies~provide an improved accuracy especially with fewer data points  (Fig.~\ref{fig:mainplots}(d)).  See Supplement~D for further experiments. 

\section{Concluding remarks}
\label{se:conclusion}

\added{
We presented Bayesian methods for discovering causal relations and for estimating linear causal effects from passively observed data. 
\gadget{} samples DAGs along a Markov chain, building on a recently introduced partition MCMC strategy \cite{Kuipers2017,Kuipers2020}, with several algorithmic modifications to improve the time and memory requirements. We have demonstrated that our method is feasible on systems with one hundred variables, and the theory (Table~\ref{table:algresults}) and simulations suggest that even larger systems, with several hundreds of variables, should be within reach. \baies{} takes as input a sample of DAGs drawn from a posterior distribution, samples model parameters conditionally on each sampled DAG to obtain a fully specified BN, thereby yielding a sampling-based approximation of the joint posterior of the causal effects; \baies{} relies on the fact that in a linear model, the effects can be efficiently computed via matrix inversion.
A similar sampling-based approach has recently been implemented also for non-linear models with binary variables \cite{Moffa17,Kuipers2019}. However, it requires either computationally expensive exact or approximate inference in the model.
%
%
Our empirical results on causal effect estimation suggest that Bayesian methods outperform non-Bayesian (IDA-based) ones especially when the data are scarce.

We conclude with two remarks. First, while our data structure for DAG sampling was motivated by a space saving, we may alternatively trade the saving for quick DAG sampling \emph{during the Markov chain simulation}. This would enable a sophisticated edge-reversal move \cite{Grzegorczyk2008}, which has proven beneficial in partition MCMC \cite{Kuipers2017} but is not implemented in \bidag{}, apparently due to its computational cost. Second, we found that \emph{optimal} sets of $K$ candidate parents often yield a good coverage of the posterior with moderate $K$, and that simple heuristics often achieve nearly optimal performance---but not always. The problem warrants further research.  E.g., could one here successfully employ techniques that quickly list large numbers of high-scoring parent sets \cite{Scanagatta2018}? We believe our approach to compare to optimal sets on moderate-size problem instances should be valuable in the quest.
}

\section*{Acknowledgments}

This work was partially supported by the Academy of Finland, Grant 316771.

\section*{Statement of broader impact}

Our work advances computational methods for learning from data. Specifically, we give more efficient and reliable methods for Bayesian statistical inference when the stucture of the underlying graphical model is unknown. A Bayesian posterior is a key enabler in informed and principled risk management and decision making under uncertainty, e.g., via the principle of expected utility; clearly, the concept of causality is essential here. We believe that, in the long run, our work will have broad impact in various areas of other sciences, technology, and in society, by making more efficient use of the available data and incorporating quantifications of uncertainty.   

Positive outcomes:
\begin{itemize}
\item This work addresses some of the key methodological challenges in computational causal inference, bringing the relatively new field closer towards high-impact real-world applications.
\item Shows the advantages of Bayesian inference, inviting and encouraging to use of similar approaches also in other domains. 
\end{itemize}

\newpage
Negative outcomes:
\begin{itemize}
\item Making causal predictions based on observational data is inherently difficult even under rather strong assumptions. Not being aware of these limitations, a non-expert user could potentially overinterpret the results.
\item Our methods contribute to the practice of discovering causal and statistical relations from data. There is a risk of biased conclusions if the data are biased (cf.\ fairness in machine learning). 
\end{itemize}

\bibliographystyle{plain}
{\small
\bibliography{paper}
}
\end{document}


\maketitle


\tableofcontents

\newpage
\appendix

\section{Algorithms}

We first recall an algorithm from the literature and then describe in detail our two novel data structures along with associated algorithms for constructing and using them. The last section describes ways to allow a node to take parents also outside the set of candidate parents. 

\subsection{Fast zeta transform}

We describe a transform that is a basic building block in several of our algorithms (see Sections~3.2 of the main article and Section~A.2 below).

Let $U = \{1, 2, \ldots, m\}$. Let $f$ be a function from the subsets of $U$ to real numbers (or, to any ring). The \emph{zeta transform} of $f$ (over the subset lattice of $U$) is the function $g$, defined for all $T \subseteq U$ by 
\bes
	g(T) := \sum_{S \subseteq T} f(S)\,.
\ees

We can evaluate the zeta transform, i.e., compute $g$ given $f$ as input, with $O(2^m m)$ additions \cite{Yates1937,Kennes1992}. This is achieved by the \emph{fast zeta transform} algorithm, which first puts $g_0 := f$ and then for $i = 1, 2, \ldots, m$ uses the recurrence 
\bes
	g_i(T) := g_{i-1}(T\setminusx\{i\}) + [i \in T]\, g_{i-1}(T)\,,
	\qquad T \subseteq U\,, 
\ees
where $[i \in T]$ evaluates to $1$ if $i$ belongs to $T$, and to $0$ otherwise. One can show that $g_m = g$.

\subsection{Preparing for catastrophic cancellations}

Lemma~1 in Section~3.2 of the main article gives us a way to compute any requested score sum by subtracting a smaller sum from a larger sum. We noted that this may result in a catastrophic cancellation due to fixed-precision arithmetic. Here we show how we handle the problematic cases by scanning through them and storing the exact (or, more accurate) values as preprocessing.

Let $i \in V$. Let $T \subseteq U \subseteq V\setminus\{i\}$. Recall that the score sum of interest is defined as  
\bes
	\tau_i(U, T) := \sum_{S \subseteq U : S \cap T \neq \emptyset} \pi_i(S)\,.
\ees
It easy to verify that, for any $j \in T$,  
\bes
	\tau_i(U, T) = \tau_i(U, \{j\}) + \tau_i(U\setminusx\{j\}, T\setminusx\{j\})\,.
\ees
In particular, this recurrence holds when $U \subseteq C_i$. Thus, for a fixed $i$, we can compute the values $\tau_i(U, T)$ for all $T \subseteq U \subseteq C_i$ with $O(3^K)$ additions. Observe that the base cases can be written as 
\bes 
	\tau_i(U, \{j\}) = \sum_{S \subseteq U \setminus\{j\}} \pi_i(S \cup \{j\})\,,
\ees
and can thus be computed using fast zeta transform with $O(2^K K^2)$ additions.

To avoid storing all the $n 3^K$ numbers, we loop over all $T \subseteq U \subseteq C_i$ and store $\tau_i(U, T)$ if and only if it cannot be reliably computed from the values $\tau_i(U)$ and $\tau_i(U\setminusx T)$, that is, if $\tau_i(U) - \tau_i(U\setminusx T) \approx 0$ (say, the relative difference is less than $2^{-32}$).

\subsection{Sampling random subsets}

Section~3.4 of the main article sketches an efficient technique for sampling a DAG from the posterior conditionally on a given root-partition. The essence of the technique is to construct a data structure for each node $i$ separately so that, given ``query'' sets $U$ and $T$, we can efficiently generate a parent set $S \subseteq U$ that intersects $T$. Below we describe our technique in more abstract terms of subset sampling.

Let $C$ be a set of $K$ elements. With each subset $X \subseteq C$ associate a weight $w(X) \geq 0$. Consider the problem of generating a random $X$ with probability proportional to $w(X)$ and satisfying $X \subseteq U$ and $X \cap T \neq \emptyset$, where $U \subseteq C$ and $T \subseteq U$ are given sets. 

We next give a data structure with the following properties:
\begin{itemize}
\item Constructing the data structure takes $O(3^K)$ time and $O(3^K)$ space.
\item Sampling a random subset takes $O(K)$ time.
\end{itemize}
For comparison, a straightforward approach would take either $O(2^{|U|})$ time (linear scan) or $O(4^K)$ space and preprocessing time (preprocessing the queries for all $T$ and $U$).

\paragraph{Construction}

For all $X \subseteq Y \subseteq C$, define 
\bes
	f(X, Y) := \sum_{X \subseteq S \subseteq Y} w(S)\,.
\ees
Observe that the function $f$ can be computed with $O(3^K)$ additions using the recurrence $f(X, Y) = f(X\cup\{i\}, Y) + f(X, Y\setminusx\{i\})$ for any $i \in Y\setminusx X$. 


\paragraph{Sampling}

Sample a subset $X$, given $U$ and $T$, as follows. For each $i \in U$ in turn, in an arbitrary order, include $i$ with probability
\be \label{eq:include}
	\frac{g(X\cup\{i\}, E)}{g(X, E)}\,,
\ee 
where $g(X, E) := f(X, U\setminusx E) - f(X, U\setminusx E \setminusx T)$ and $X$ and $E$ denote the sets of elements that were already included and excluded, respectively; initially, we set both $X$ and $E$ empty. 

To see that $X$ is generated with the correct probability, observe first that the probability of excluding $i$ can be written as 
\be \label{eq:exclude}
	\frac{g(X, E\cup\{i\})}{g(X, E)}\,.
\ee
Namely, a set $S$ contributes to the denominator with weight $w(S)$ exactly when $X \subseteq S \subseteq U\setminusx E$ and $S \cap T \neq \emptyset$, and to the numerator in \eqref{eq:exclude} exactly when, in addition, $i \not\in S$, and to the numerator  in \eqref{eq:include} exactly when, in addition, $i \in S$. 
 
Thus, the probability of the decision made in the $t$th round is $g(X_t, E_t)/g(X_{t-1}, E_{t-1})$, where $X_t$ and $E_t$ are the sets of elements included and excluded after the first $t$ rounds. By the chain rule, we get from the telescoping product that $X := X_{|U|}$ is generated with probability $g(X, U\setminusx X)/g(\emptyset, \emptyset)$. Observe that $g(X, U\setminusx X) = w(X)$ if $X$ intersects $T$, and $g(X, U\setminusx X) = 0$ otherwise. 

\paragraph{Confronting catastrophic cancellation}

Due to fixed-precision arithmetic, the computed value of $g(X, E)$ may be zero even if the exact value was non-zero. This approximation may result in an uncontrolled bias in the sampling distribution. 

As a remedy, if the computed value $g(X, E)$ has a large relative error (deduced by the terms in the subtraction), we switch over to brute-force sampling, which takes time $O(2^{|U|})$. 

\subsection{Allowing parents outside the candidates}

Strictly requiring all parents of each node $i$ to come from the established set of $K$ candidate parents $C_i$ has two drawbacks: (i) For some nodes $i$, the found set $C_i$ may not be optimal or sufficiently large to cover the posterior well. (ii) The posterior landscape may contain large zero-probability regions, which makes moving between node partitions inefficient for the Markov chain. A remedy for both issues is to allow any single node $j \neq i$ be a parent of $i$, either in combination with a small number of other arbitrary nodes, like implemented in \gadget{}, or in combination with any number of other parents from the candidates, like implemented in \bidag{} \cite{Kuipers2020}. Below we present some further details.  

\paragraph{Implementation in \gadget{}}

Currently \gadget{} allows a node $i$ to have a parent set that either is contained in the set of candidates $C_i$, or is of size at most $d$, where $d$ is a user parameter, set to $3$ in our experiments. Compared to the basic case of $d = 0$, this extension requires some additional work both in preprocessing and in the Markov chain simulation phase. 

In preprocessing, we compute the local score for $O(n^d)$ parent sets per node, in addition to the $O(2^K)$ subsets of the candidate set. Adopting previously proposed techniques \cite{Friedman2003,Niinimaki2013}, we sort the parents sets in decreasing order by the score; this will enable a tolerably fast computation of any queried partial sum of the scores to within a given relative error.   

In the simulation phase, when the score sum is needed for a node $i$ over the parent sets that are contained in $U$ and intersect $T$, we scan the sorted list until the accumulated sum is quaranteed to be sufficiently large (we allowed a relative error of $0.1$ in our experiments). Compared to previous implementations of this idea \cite{Friedman2003,Niinimaki2013}, a distinctive feature in our implementation is that we can initiate the accumulating sum by the partial sum contributed by the parent sets that are contained in the candidates; this contribution is often non-zero and expedites the computation.

\paragraph{Implementation in \bidag{}} 

Constant-time score sum computation during simulation is maintained in \bidag{} by precomputing the score sums for the extended parent sets. This increases the space requirement and the preprocessing time by a factor of $n$, further increasing the gap to the complexity bounds of \gadget{}, necessitating the use of a smaller value of $K$.  

Furthermore, a preliminary simulation run is used to extend the initial candidate parent sets (found by the PC algorithm) based on visited high-scoring DAGs \cite{Kuipers2020}. The extension is vital for this procedure.

\section{Bayesian posterior: a derivation}


Here we derive  formulas for the posterior of the parameters of a linear Gaussian DAG model assuming a normal--Wishart prior. We follow previous similar derivations by Geiger and Heckerman \cite{Geiger02} and Kuipers, Moffa, and Heckerman \cite{kuipers2014}.
 We refer to the two reference articles by GH and KMH. In more detail, GH derive a prior distribution for the model parameters similar to ours.  KMH derives the BGe score, noting  errors in the original derivation~\cite{gh}. We derive here the prior and the posterior of the model parameters, taking into account the KMH corrections and further correcting additional inconsistencies in GH.

\subsection{Prior and posterior distributions with respect to all variables}

The basic idea is to first consider a complete DAG. We will specify the prior so that it does not distinguish between equivalent DAGs. Thus, it does not matter which complete DAG we consider. Then, when we proceed to consider a node $i$ in an arbitrary DAG, we can make use the result we have for any complete DAG that contains the same local pattern, i.e., node $i$ has the same parent set.

We assume $\x$ is distributed normally with precision matrix $W$ and mean $\muu$:
\bes
	\x \sim \mathcal{N} (\muu,W)\,.
\ees
Following GH and KMH, $\muu$ and $W$ have a (conjugate) normal--Wishart prior distribution:
\bes
	\muu \sim \mathcal{N}(   \nuu, \alpha_{\mu} W  )\,, 
	\qquad 
	W \sim \mathcal{W}_n(  T^{-1},\alpha_{w})
\ees
where $\alpha_{\mu}$ and $\alpha_w$ are equivalent sample sizes, $\nuu$ is a mean vector,  and $T$ is an inverse scale matrix. 
Because this is a conjugate prior to normal likelihood,  we have that the posterior is of the same form:
\bes
	\muu \sim \mathcal{N}(   \nuu', \alpha'_{\mu} W  )\,, 
	\qquad 
	W \sim \mathcal{W}_n(  R^{-1},\alpha'_{w})
\ees
with updated hyperparameters $\alpha_\mu' := \alpha_\mu+N$, $\alpha'_{w} :=  \alpha_{w}+N$,  
\bes
	\nuu' := \frac{\alpha_\mu\nuu+N\bar{\x}_N}{\alpha_\mu+N}\,, 
	\quad 
	\textrm{and}
	\quad 
	R := T +S_N+\frac{\alpha_\mu N}{\alpha_\mu+N}(\nuu - \bar{\x}_N)(\nuu - \bar{\x}_N){\tran}\,,
\ees
where $ \bar{\x}_N= \frac{1}{N}\sum_{s=1}^N \x_s$ and $S_N=\sum_{s=1}^N (\x_s-\bar{\x}_N)(\x_s-\bar{\x}_N){\tran}$.

\subsection{Prior and posterior distributions with respect to subsets of variables}

Now consider a node $i$ in an arbitrary DAG. Our goal is to determine the joint posterior distribution of the coefficients $\b_i$ associated with the edges from $pa(i)$ to $i$ and the precision of the error term $\precision_i$. Under the modularity assumption, the local distribution and parameter prior of node $i$ is the same for all DAGs where $pa(i)$ is the parent set of node $i$. In particular, let us consider a complete DAG that has this property and where, in addition, the topological ordering within $\pa(i)$ conincides with the natural ordering of integers. Let $Y := \pa(i)\cup \{i\}$, let $l$ be the size of $Y$, and let $\sY$ denote the set of remaining $n-l$ nodes. We will need the fact that subgraph induced by $Y$ is complete in the change of parameterization later on.  
For a vector $\vect{v}$ and set $S$, we let $\vect{v}_S$ denote the subvector $(v_j : j \in S)$ where we order the entries in increasing order by $j$, except that $i$ is the last if it belongs to $S$. We use an equivalent notation for submatrices indexed by subsets of rows and columns. 


Following KMH, we consider the subvector $\y=\x_\Y$ which has the distribution
\bes
	\y \sim \mathcal{N} (\muu_{\Y},W_{\Y})
\ees
where $W_{\Y} := W_{\Y\Y}-W_{\Y \sY}(W_{\sY \sY})^{-1} W_{\sY \Y}$ is obtained by inverting to covariance matrix, marginalizing and inverting back to precision. 
The prior on $\muu_{\Y}$ and $W_{\Y}$ is 
\bes
	\muu_{\Y} \sim \mathcal{N}( \nuu_\Y,\alpha_{\mu}W_{\Y})\,,
	\qquad 
	W_{\Y} \sim \mathcal{W}_l\left((\TY)^{-1},\alpha_w-n+l\right)\,.
\ees
This result is obtained in Equation~A.24 in KMH. To get to the posterior, we can make a similar transformation of the full posterior to this subset:
\bes
	\muu_{\Y} \sim \mathcal{N}( \nuu'_\Y,\alpha'_{\mu}W_{\Y})\,,
	\qquad 
	W_{\Y} \sim \mathcal{W}_l\left((\RY)^{-1},\alpha_w'-n+l\right)\,.
\ees
Note that the degrees of freedom in the above Wishart distribution has been reduced compared to the corresponding distribution in the previous subsection.
This result is in Equation~A.26 in KMH.


\subsection{Change of parameterization} 

We have that $\BY$ is a full lower triangular $l \times l$ matrix with $(\b_i,0)$ as the $l$th row. Likewise, $\precisionmat_{\Y \Y}$ is an $l\times l$ diagonal matrix including the precisions of the error terms with $\precision_i$ as the last element.  We will utilize the structural equation model
\bes
	\y := \muu_\Y + \BY (\y - \muu_\Y) + \e_\Y\,,
\ees
from which we can solve for $\y$:
\bes
	\y = \muu_\Y  +(I - \BY)^{-1} \e_\Y\,.
\ees
%
%
%
Now we change the parameterization from $W_\Y$  to  $(\BY, \precisionmatY)$ using a bijective transformation $f$. Following Gelman et al.\ \cite[p.~21--22]{gelman2013bayesian}, the density function of $(\BY, \precisionmatY)$ is obtained as 
\bes
	p\big(\BY,\precisionmatY\big) = |\det J|
	\cdot p\big(f^{-1}(\BY, \precisionmatY)\big) 
\ees
where 
\bes
	f^{-1}(\BY,\precisionmatY) =  (I - \BY)^{\ptran} \precisionmatY (I - \BY),
\ees
and $J$ is the Jacobian matrix, i.e., the square matrix of partial derivates of $f^{-1}$; $f$ is one-to-one, since $\BY$ corresponds to a full DAG. The matrix $W_\Y$ can be represented in a block form as follows: Denote by $B_{11}$ (resp.\ $\precisionmat_{11}$) the submatrix of $\BY$  (resp.\ $\precisionmatY$) where the last row and the last column are removed. Now we have 
\bes
	W_\Y &=& 
		\left[\begin{array}{ccc}
		(I-B_{11}){\tran} & -\b_i \\
		0 & 1
		\end{array} \right] \left[\begin{array}{ccc}
		\precisionmat_{11} & 0 \\
		0 & \precision_i
		\end{array} \right]
		\left[\begin{array}{ccc}
		I-B_{11} & 0 \\
		-\b_i{\tran} & 1
		\end{array} \right] \\
	&=&
		\left[\begin{array}{ccc}
		(I-B_{11}){\tran}\precisionmat_{11} &  -\b_i  \precision_i\\
		0 & \precision_i
		\end{array} \right]
		\left[\begin{array}{ccc}
		I-B_{11} & 0 \\
		-\b_i{\tran} & 1
		\end{array} \right]  \\
	&=&
		\left[\begin{array}{ccc}
		(I-B_{11}){\tran}\precisionmat_{11}(I-B_{11})+\precision_i \b_i\b_i{\tran} & 			-\precision_i\b_i \\
		-\precision_i\b_i{\tran} & \precision_i
		\end{array} \right]\,.
\ees

The absolute value of the Jacobian determinant can be obtained by direct calculation using a similar recursion as in Theorem~6 of Geiger and Heckerman~\cite{gh}:
\begin{eqnarray}
		|\det J | = \prod_{j\in \Y}\precision_j^{k_j-1} 
	\label{eq:jacobian}
\end{eqnarray}
where $k_j$ is the index of the node $j$ in $Y$. Note that this product contains the factor $q_i^{l-1}$.

\subsection{Posterior of the coefficients and the precision}

Following KMH, the posterior density function of the $k$-dimensional Wishart distribution is 
\begin{eqnarray}
	\mathcal{W}_k(W|T^{-1},\, \alpha_w ) &=& \frac{|W|^{(\alpha_w-k-1)/2}}{Z_W(k,T,\alpha_w)} \exp\Big\{-\frac{1}{2}\tr(TW)\Big\}\,, \label{eq:wishart}  
\end{eqnarray}
where $Z_W(k,T,\alpha_w)$ is the normalizing constant and $W$ is positive definite.
Plugging in the parameters $T := \RY$, $W := W_\Y$, $k := l$, and $\alpha_w := \alpha_w'-n+l$ yields
\bes
	\mathcal{W}_l\left(W_\Y|(\RY)^{-1},\; \alpha_w'-n+l \right) &=& 
\frac{|W_\Y|^{(\alpha_w'-n-1)/2}}{Z_W(l, \RY,\alpha_w'-n+l)} \exp\Big\{-\frac{1}{2}\tr(\RY W_\Y)\Big\}\,.
\ees
The change of parametrization then gives us the posterior density 
\bes
	\pi(\BY, \precisionmatY) &\propto& \Big(\prod_{j\in \Y}\precision_j^{k_j-1}\Big)\big|(I - \BY)\tran 	\precisionmatY (I - \BY)\big|^{(\alpha_w'-n-1)/2} \\
	& & \qquad \times \exp\Big\{-\frac{1}{2}\tr\big(\RY(I - \BY)\tran \precisionmatY (I - \BY)\big)\Big\}\,.
\ees

The trace term in the exponent can be calculated by blocks: 
\begin{eqnarray}
	&& \tr\big(\RY (I - \BY)\tran \precisionmatY (I - \BY)\big) \nonumber\\
	&=&
 	\tr\Bigg(\left[\begin{array}{cc}
		R_{11} & R_{12}\\
		R_{21} &R_{22}
		\end{array}\right]\left[\begin{array}{cc}
		(I-B_{11})\tran\precisionmat_{11}(I-B_{11})+q_i\b_{i}\b_{i}	{\tran} & -\precision_i\b_{i}\\
		-\precision_i\b_{i}{\tran} & \precision_i
		\end{array}\right]\Bigg) \nonumber\\
	&=&
 	\tr\big(R_{11}(I-B_{11})\tran\precisionmat_{11}(I-B_{11})+\precision_iR_{11}\b_{i}\b_{i}{\tran} -  \precision_iR_{12}\b_{i}{\tran}\big)+ R_{22}\precision_i-\precision_iR_{21}\b_{i} \nonumber\\
	&=& \precision_i\b_{i}{\tran}R_{11}\b_{i} -2\precision_iR_{21}\b_{i} +  R_{22}\precision_i+ c \nonumber\\
	&=& \precision_i \b_{i}{\tran}R_{11}\b_i   - 2\precision_iR_{21}(R_{11})^{-\ptran}R_{11} \b_i + \precision_i R_{21}(R_{11})^{-\ptran}R_{11}(R_{11})^{-1}R_{12}\nonumber\\
 	& & \qquad -\;\precision_i R_{21}(R_{11})^{-\ptran}R_{11}(R_{11})^{-1}R_{12}+  R_{22}\precision_i+ c \nonumber\\
	&=& \precision_i \big(\b_{i} - (R_{11})^{-1}R_{12}\big){\tran} R_{11} \big(\b_{i}-(R_{11})^{-1} R_{12}\big) +\; \precision_i \big(R_{22}- R_{21}(R_{11})^{-1}R_{12}\big) + c\,, \label{eq:exp}
\end{eqnarray}
where $c$ collects any terms that are constant with respect to $\b_{i}$ and $\precision_i$. 

The determinant term simplifies since the determinant of a triangular matrix is the product of its diagonal entries:
\begin{eqnarray}
	\big|(I - \BY)^{\ptran} \precisionmatY (I - \BY)\big|^{(\alpha_w'-n-1)/2} 
	=
	|\precisionmatY|^{(\alpha_w'-n-1)/2} 
= \prod_{j\in \Y} \precision_j^{(\alpha_w'-n-1)/2}\,.  \label{eq:det}
\end{eqnarray}

Putting together the exponent (Eq.~\ref{eq:exp}), determinant (Eq.~\ref{eq:det}), and the Jacobian  (Eq.~\ref{eq:jacobian}) gives
\be
	\pi(\b_{i},\precision_i) 
	\,\propto\, 
	\precision_i^{l-1} \precision_i^{(\alpha_w'-n-1)/2} 
	\exp\Big\{ -\frac{1}{2} 
	\Big[\precision_i \big(\b_{i} - (R_{11})^{-1}R_{12}\big){\tran} R_{11} \big(\b_{i}-(R_{11})^{-1} R_{12}\big) \nonumber \\
	+\; \precision_i \big(R_{22} - R_{21}(R_{11})^{-1} R_{12} \big) \Big] \Big\}\,. \label{eq:joint}
\ee
The first term in the exponent implies that:
\bes
	\b_i \;|\; \precision_i \,\sim\, \mathcal{N}\left((R_{11})^{-1}R_{12}\,,\;\precision_iR_{11}\right)\,.
\ees

The normalizing constant for the normal distribution includes the term 
\bes
	\big|(\precision_iR_{11})^{-1} \big|^{-1/2} \,=\, \big|(\precision_iR_{11})\big|^{1/2} \,\propto\, \precision_i^{(l-1)/2}\,. 
\ees
Thus, marginalizing out $\b_i$ leaves
\bes
	\pi(\precision_i) 
	\,\propto\, 
	\precision_i^{(l-1)/2} \precision_i^{(\alpha_w'-n-1)/2} 
	\exp\Big\{ -\frac{1}{2} \precision_i \big(R_{22}- R_{21}(R_{11})^{-1} R_{12} \big) \Big\}\,.
\ees
This is a one-dimensional Wishart (Gamma) distribution, see Equation~\ref{eq:wishart}. Thus
\bes
	\precision_i \sim \mathcal{W}_1 \Big(  \big(R_{22}- R_{21}(R_{11})^{-1} R_{12}\big)^{-1}\,,\; \alpha_w'-n+l  \Big)\,.
\ees

\subsection{Prior of the coefficients and the precision}

If we replace $R$ with $T$ and $\alpha'_w$ with $\alpha_w$ in the above derivation, we can obtain the priors:
\bes
\b_i \;|\; \precision_i \sim \mathcal{N}\Big((T_{11})^{-1}T_{12}\,,\; \precision_i T_{11}\Big)\,, 
	\qquad
	\precision_i \sim \mathcal{W}_1 \Big(  \big(T_{22}- T_{21}(T_{11})^{-1}T_{12} \big)^{-1}\,,\; \alpha_w-n+l  \Big)\,.	
\ees
Compared to p.~1425 in GH the precision/covariance of $\b_i$ is different. The dimensions obtained in our derivation correspond to the dimensions of $\b_i$ correctly. Furthermore, the degrees of freedom differ; ours take into account the change due to considering subset of variables, pointed out by KMH.

\subsection{Marginal posterior of the edge coefficients} 
We can still integrate out $\precision_i$ to get the marginal posterior of $\b_i$. The non-constant terms in the joint density in Equation~\ref{eq:joint} are 
\bes
	\precision_i^{(\alpha_w'-n+2l-3)/2} 
	\exp\Big\{
	 -\frac{1}{2}\Big[ \big(\b_{i}-  (R_{11})^{-1}R_{12}\big){\tran} R_{11} \big(\b_{i}-(R_{11})^{-1} R_{12}\big)\\
	+\; \big(R_{22}- R_{21}(R_{11})^{-1} R_{12} \big)\Big]\precision_i\Big\}\,.
\ees
Integrating this over $\precision_i$ results in a Gamma integral, which evaluates to
\bes
	\Gamma\big((\alpha_w'-n+2l-1)/2\big) 
	\,\Big\{ \frac{1}{2}\Big[ \big(\b_{i}-  (R_{11})^{-1}R_{12}){\tran} R_{11} \big(\b_{i}-(R_{11})^{-1}R_{12}\big)\\ 
	+\; \big(R_{22}- R_{21}(R_{11})^{-1}R_{12} \big)  \Big]  
	\Big\}^{-(\alpha_w'-n+2l-1)/2}\,.
\ees
This implies that
\bes
	\pi(\b_i) 
	\,\propto\, \bigg( 
	1+\frac{1}{\alpha_w'-n+l} \; 
	\big(\b_{i} - (R_{11})^{-1}R_{12}\big){\tran}\;
	\frac{\alpha_w'-n+l}{R_{22} - R_{21}(R_{11})^{-1} R_{12}}\;\\
	\times\; R_{11} \big(\b_{i} - (R_{11})^{-1} R_{12}\big)  
	\bigg)^{-(\alpha_w'-n+l+l-1)/2}\,,
\ees
and since $\b_i$ has $l-1$ elements, we have that (see, e.g., Gelman et al.~\cite{gelman2013bayesian})  
\bes
	\b_i \,\sim\, t_{l-1} \bigg( (R_{11})^{-1}R_{12}\,,\;   
	\frac{\alpha_w'-n+l}{R_{22}- R_{21}\,(R_{11})^{-1} R_{12}} R_{11}\,,\;
	\alpha_w'-n+l \bigg)\,,
\ees
where the middle term marks precision.


\section{Candidate parent selection}

Here we describe in detail the algorithms used for selecting the $K$ candidate parents and how the performance of the algorithms was evaluated. Some implementation practicalities are also discussed.

\subsection{Optimal algorithm and heuristics}




Unless otherwise specified, the local scores referred to in the following are as specified in section~D.1.

\begin{itemize}[leftmargin=64pt]
\item[\emph{Opt}] The \emph{Opt} (i.e., optimal) algorithm selects a $K$-set $C_i$ so as to maximize the posterior probability that $\pa(i) \subseteq C_i$ (cf.\ Proposition~2 in the main paper). First all the local scores are computed, after which the (unnormalized) parent set probabilities are computed by summing for each node and parent set the scores of DAGs where the variable has the given parent set. As a last step, for each node $i$ and subset of nodes $C_i \subseteq V \setminus \{i\}$ of size $K$ (that is, for each possible set of candidate parents) the sum of probabilities over the subsets of $C_i$ is computed, and the set maximizing the sum is finally output. Note that \emph{Opt} is scalable only up to around 25 variables and we use it here for a reference.

\item[\emph{Top}] Select the $K$ nodes $j$ with the highest local score $\pi_i(\{j\})$. The heuristic therefore only considers parent sets of size $1$ and will miss any candidate parent whose value is dependent on being in a set with a number of others. 

\item[\emph{PCb}] Merge the neighbourhoods of $i$, excluding children, returned by \emph{PC} on $20$ bootstrap samples. The parameters of the algorithm are the p-value threshold and the maximum conditioning set size in the independence tests. To avoid being overly conservative, we set the p-value to $0.10$ and the maximum conditioning set size to 1.
    
\item[\emph{MBb}] Merge the Markov blankets of $i$ returned by the incremental association (\emph{IA}) algorithm on $20$ bootstrap samples. The p-value threshold and maximum conditioning set size were set to the same values as in \emph{PCb}.
  
\item[\emph{GESb}] Merge the neighborhoods of $i$, excluding children, returned by greedy equivalence search (\emph{GES}) on $20$ bootstrap samples. The score function used (in pcalg \cite{kalisch:2012, hauser:2012}) is BIC.

\item[\emph{Greedy}] Iteratively, add a best node to $C_i$, initially empty, where \emph{goodness} of $j$ is $\max_{S \subseteq C_i} \pi_i(S\cup\{j\})$. That is, add the node with which we can get the next highest uncovered local score covered.

\item[\emph{Back\&Forth}] Using the definition of goodness from \emph{Greedy}, start from a random $K$-set, delete a worst and add a best node, alternatingly, until the added node is the one deleted in the previous step.

\item[\emph{Greedy-lite}] A computationally more efficient variant of \emph{Greedy}. First, build a candidate set $C_i$ of $K-s$ nodes with \emph{Greedy}. Then, instead of adding the single best node, add the $s$ best nodes in a single step, where the goodness of a node is defined as in \emph{Greedy}. We set $s := 6$ to limit the number of scores that we have to compute by a factor of $2^6 = 64$, as compared to \emph{Greedy}.
\end{itemize}

\emph{Gadget}, including the candidate selection phase, in its current version is implemented mostly in Python, with some time critical parts in C++. The local scores are computed with the Python version of Gobnilp \cite{Bartlett2017,cussens:2011}. The PC and Incremental Association algorithms are implemented in the bnlearn R-package \cite{scutari:2010}, which our code interfaces with. Similarly, we use GES as implemented in the pcalg R-package \cite{kalisch:2012, hauser:2012}. To compute the marginal posterior parent set probabilities (as per Proposition~2 in the main paper), allowing both for the evaluation of the heuristics and for computing the optimal parent sets when $n$ is small, we use software developed by Pensar et al. \cite{Pensar2020}.    

As the algorithms \emph{PCb}, \emph{MBb} and \emph{GESb} as described can return any number of candidate parents for each node, there has to be a mechanism for adjusting the number to match the desired $K$ exactly. In our experiments we tried two approaches: adding (removing) nodes randomly, or in the order given by the scores of their singular parent sets. The latter proved more performant and was therefore used. Consequently, the selection of the parameters for the used PC and IA algorithms also determines how closely the returned candidates mirror those of \emph{Top} -- if the initial phase of the heuristics return an empty graph, the candidates finally returned equal those given by \emph{Top}. Bootstrapping the input data has a similar effect, as it can only increase the number of candidates returned by the initial parts of the heuristics. 

In terms of speed, the scoring code we use does not seem particularly well suited to \emph{Opt}, or the heuristics \emph{Greedy} and \emph{Back\&Forth}, which require large numbers of scores to be computed. On the other hand, \emph{Greedy} seems like a good candidate as the default algorithm, as it achieves the greatest coverage of the posterior mass for sufficiently large $K$, close to that of the reference \emph{Opt}. \emph{Back\&Forth} possibly offers only a slight advantage over \emph{Greedy} in some cases (Figure~2 in main paper; Figure~\ref{fig:coverage}). Thus, in order to avoid the candidate selection phase dominating the time use in the MCMC runs, while still constructing candidate sets that cover close to equal amount of the posterior mass as those constructed by \emph{Greedy}, we used the more efficient \emph{Greedy-lite} variant of it for the main experiments in the paper.

Apart from the heuristics listed, we also experimented with numerous others. These included, for example, hybrid ones which ran a number of heuristics in parallel for increasing $K' \leq K$ until the size of the union of the candidates they found reached the target $K$. The results, however, did not show marked improvement over the simpler methods.

\subsection{Test data}

Gaussian data was generated as explained in section~4.2 of the main paper.

For the experiments on discrete data, we used the UCI data sets utilized by Malone et al.~\cite{Malone:2018} for learning discrete Bayesian networks. 
In the paper we included all the data sets with up to 23 variables, to allow for exact evaluation of the parent set probabilities\footnote{We did not include {\sc Letter} ($n = 17$, $N = 20\, 000$), however, which proved to be too difficult to compute all local scores for with our setup, presumably due to the large number of data points.}.


\subsection{Empirical results}

\begin{figure}[t!]	
  \centering

\hspace*{-15pt}
  \begin{subfigure}[t]{.45\textwidth}
    \includegraphics[scale=0.475]{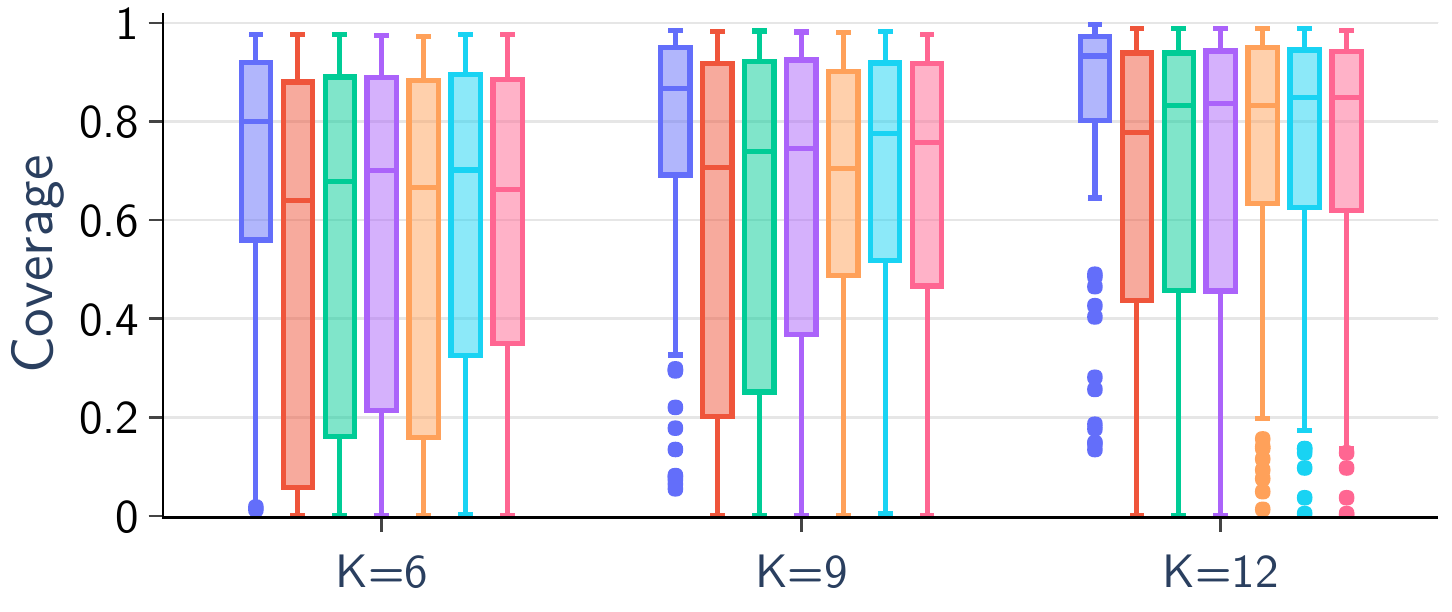}
    \caption{Gaussian, $n = 20$, $N = 50$}\label{fig:coverage:zoo}
  \end{subfigure}
\hspace*{20pt}
  \begin{subfigure}[t]{.45\textwidth}
    \includegraphics[scale=0.475]{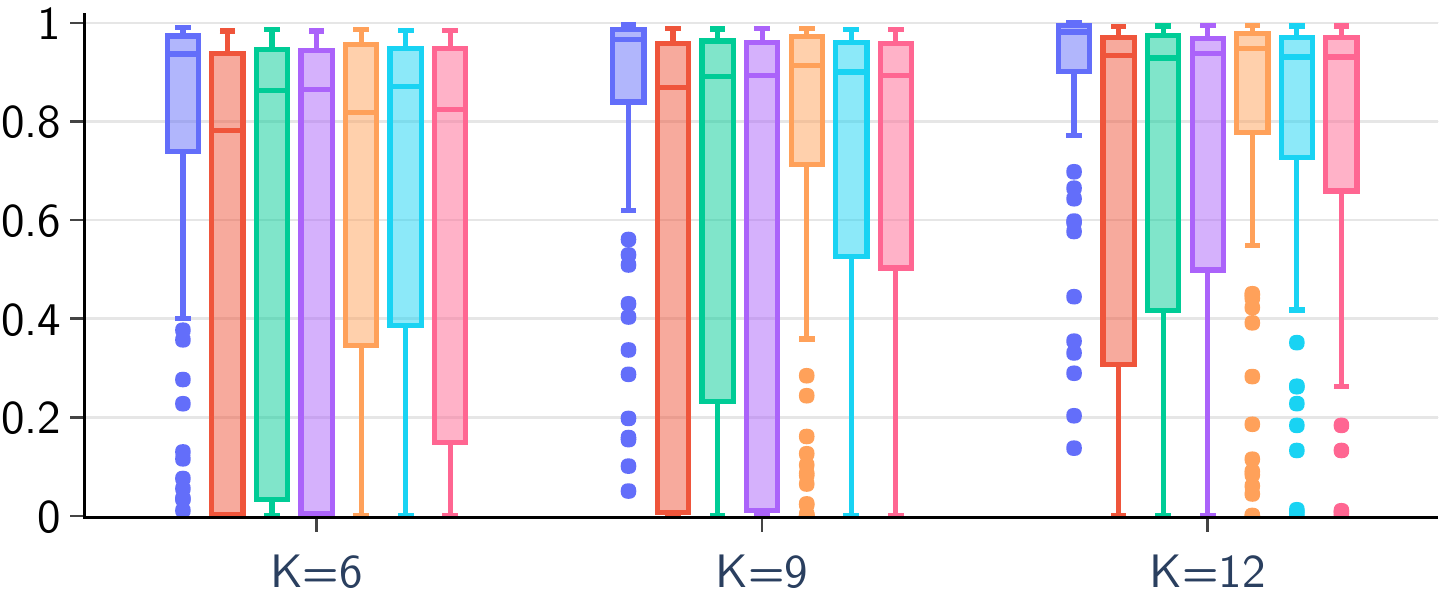}
    \caption{Gaussian, $n = 20$, $N = 200$}\label{fig:coverage:lymph}
  \end{subfigure}  
  \medskip

  \begin{subfigure}[t]{1.0\textwidth}
    \centering
    \includegraphics[scale=0.675]{figures/legend-gaussian-heuristics.pdf}
  \end{subfigure}  
  
  \caption{Distribution of coverages over the $n$ nodes for 4 randomly selected Gaussian datasets.}
\label{fig:coverage-gaussian}
\end{figure}

\begin{figure}[t!]	
  \centering

\hspace*{-4pt}
  \begin{subfigure}[t]{.30\textwidth}
    \includegraphics[scale=0.475]{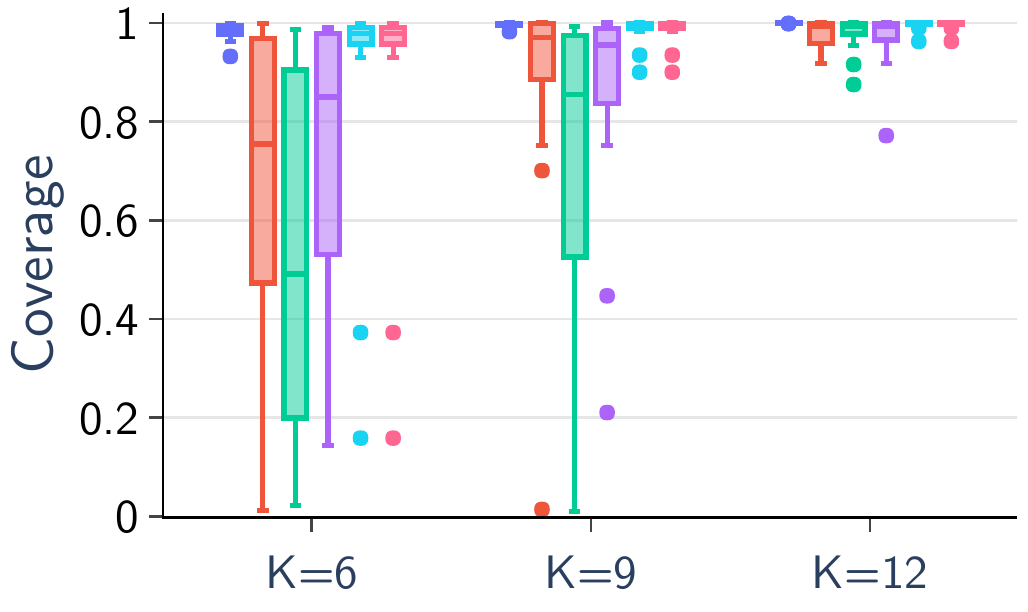}
    \smallcaption{{\sc Voting}, $n = 17$, $N = 435$}\label{fig:coverage:voting}		
  \end{subfigure}
\hspace*{17pt}
  \begin{subfigure}[t]{.30\textwidth}
    \includegraphics[scale=0.475]{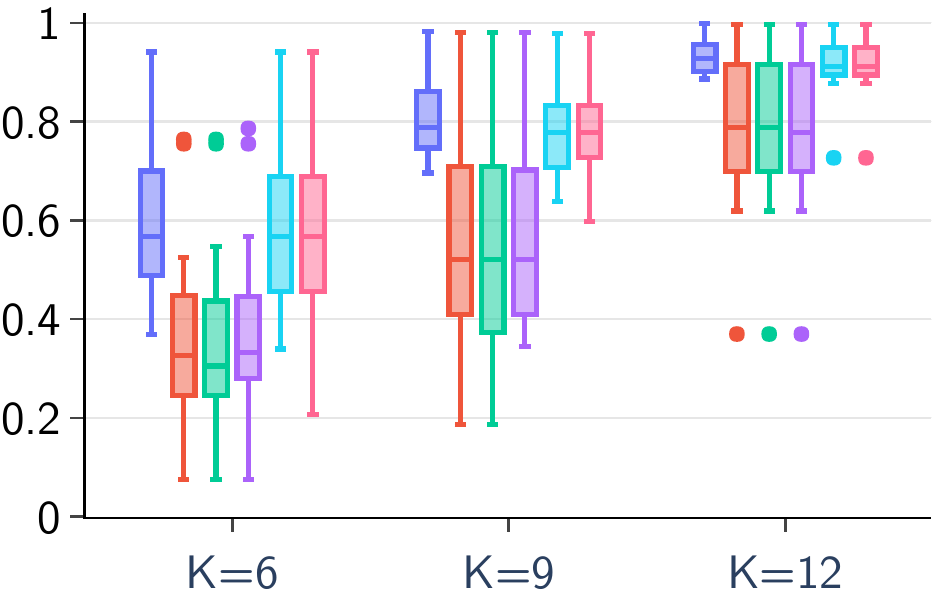}
    \smallcaption{{\sc Zoo}, $n = 17$, $N = 101$}\label{fig:coverage:zoo}
  \end{subfigure}
\hspace*{7pt}
  \begin{subfigure}[t]{.30\textwidth}
    \includegraphics[scale=0.475]{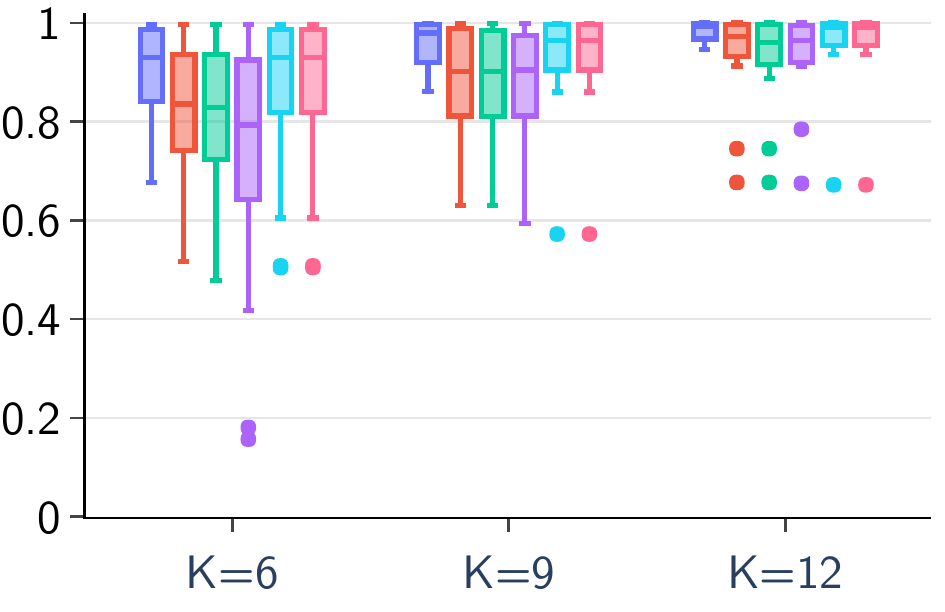}
    \smallcaption{{\sc Lymph}, $n = 18$, $N = 148$}\label{fig:coverage:lymph}
  \end{subfigure}  
  \medskip

\hspace*{-4pt}
  \begin{subfigure}[t]{.30\textwidth}
    \includegraphics[scale=0.475]{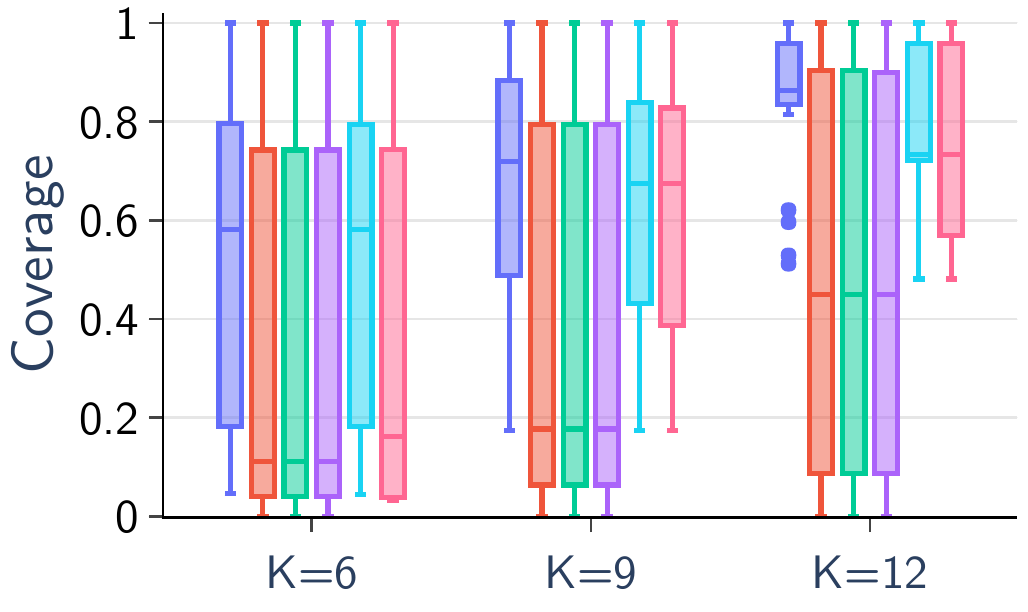}
    \smallcaption{{\sc Eucalyptus}, $n = 20$, $N = 736$}\label{fig:coverage:eucalyptus}		
  \end{subfigure}
\hspace*{17pt}
  \begin{subfigure}[t]{.30\textwidth}
    \includegraphics[scale=0.475]{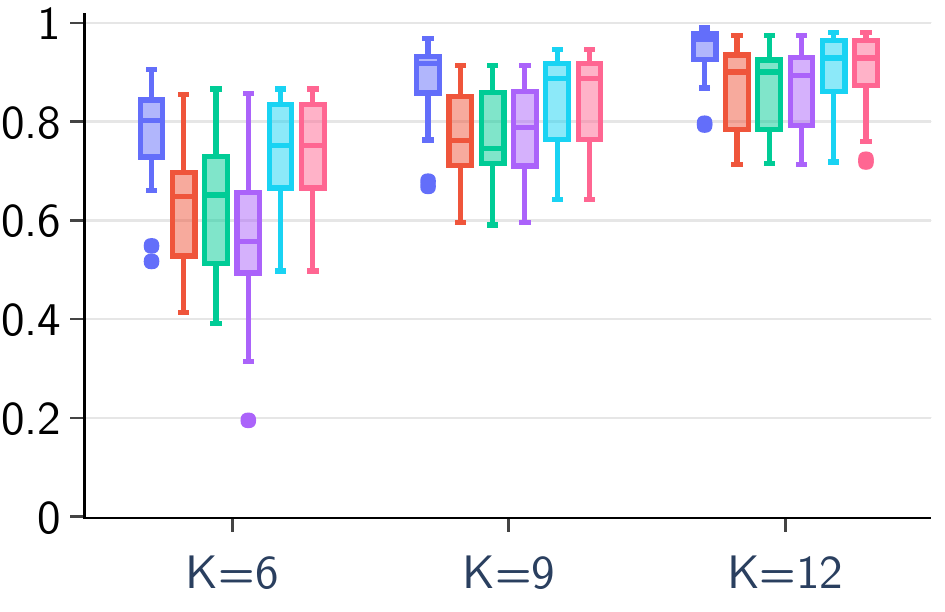}
    \smallcaption{{\sc Hepatitis}, $n = 20$, $N = 155$}\label{fig:coverage:hepatitis}
  \end{subfigure}
\hspace*{7pt}
  \begin{subfigure}[t]{.30\textwidth}
    \includegraphics[scale=0.475]{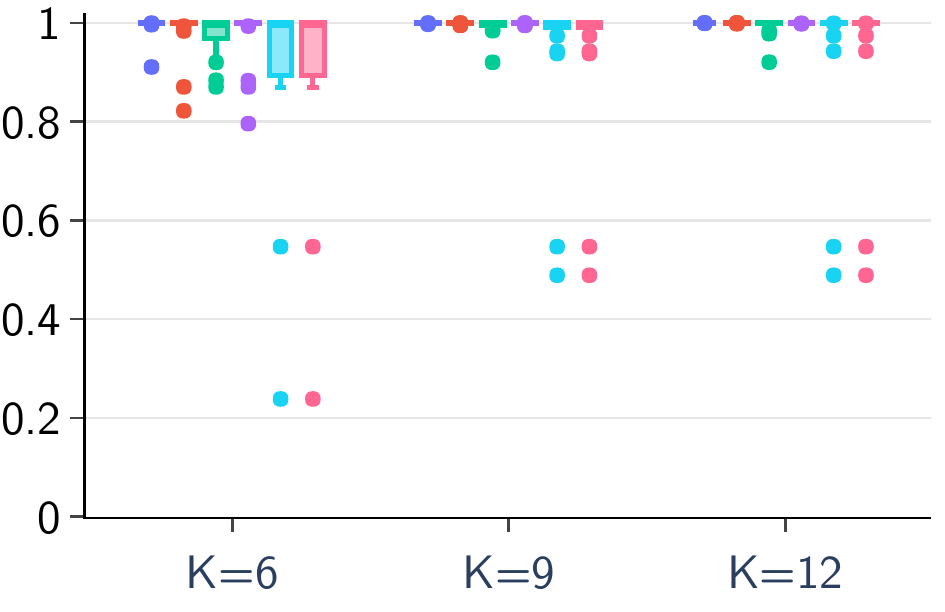}
    \smallcaption{{\sc Credit-g}, $n = 21$, $N = 1000$}\label{fig:coverage:credit-g}
  \end{subfigure}  
  \medskip

\hspace*{-4pt}
  \begin{subfigure}[t]{.30\textwidth}
    \includegraphics[scale=0.475]{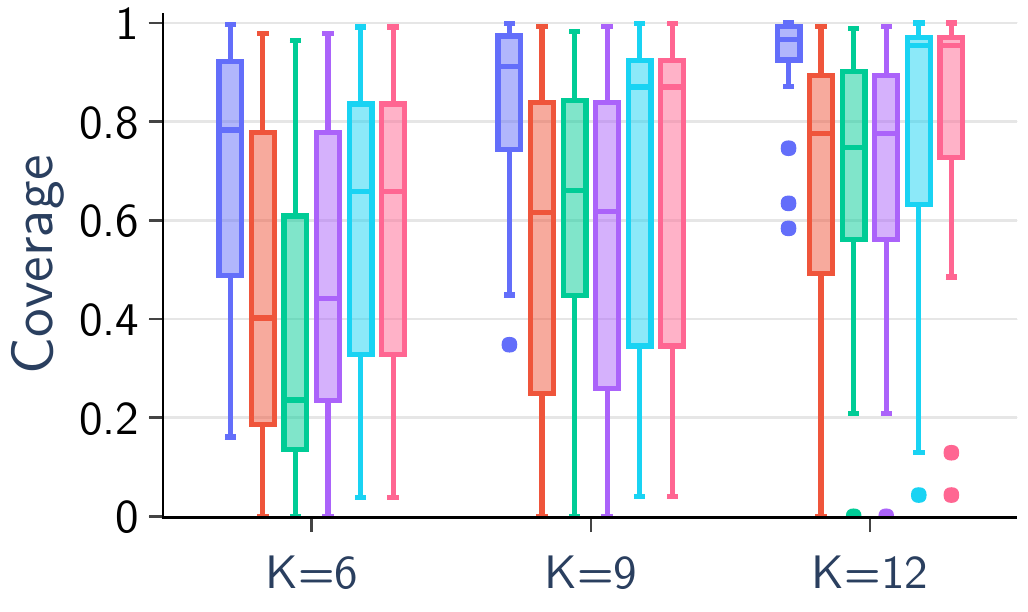}
    \smallcaption{{\sc Hypothyroid}, $n = 22$, $N = 3772$}\label{fig:coverage:hypothyroid}		
  \end{subfigure}
\hspace*{17pt}
  \begin{subfigure}[t]{.30\textwidth}
    \includegraphics[scale=0.475]{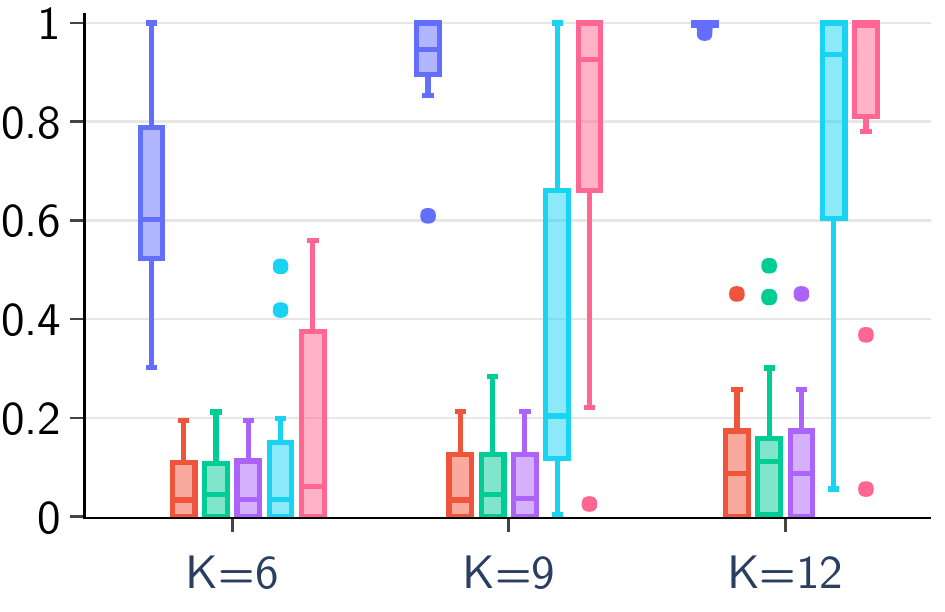}
    \smallcaption{{\sc Mushroom}, $n = 22$, $N = 8124$}\label{fig:coverage:mushroom}
  \end{subfigure}
\hspace*{7pt}
  \begin{subfigure}[t]{.30\textwidth}
    \includegraphics[scale=0.475]{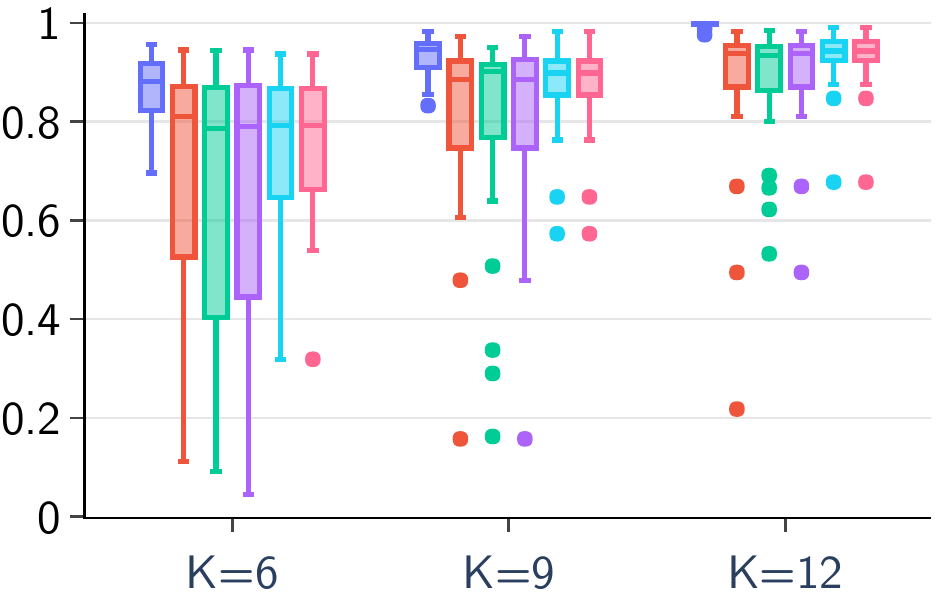}
    \smallcaption{{\sc Spect}, $n = 23$, $N = 267$}\label{fig:coverage:spect}
  \end{subfigure}  
  \medskip

  \begin{subfigure}[t]{1.0\textwidth}
    \centering
    \includegraphics[scale=0.675]{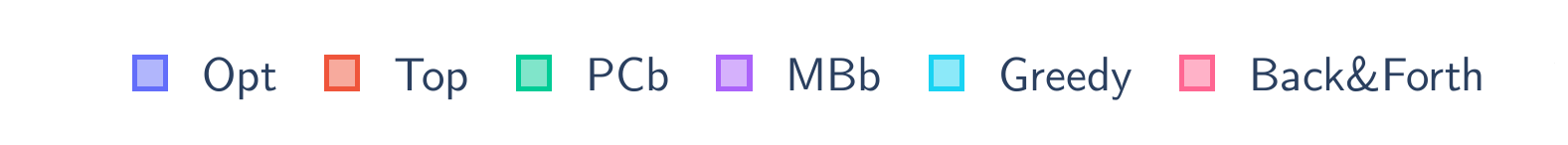}
  \end{subfigure}  
  
  \caption{Distribution of coverages over the $n$ nodes for each included UCI data set.}
\label{fig:coverage}
\end{figure}

To evaluate the returned candidates for a given node, when the number of variables is sufficiently small to allow for computing the exact parent set posteriors, we simply summed over all the posteriors of the subsets of the candidates. Finally, we reported the mean over the nodes of the \emph{coverages} thus obtained (Figure~2 in the main paper). Here we break down the analysis further by considering the distribution of the posterior mass covered by the candidate parents of each node. The results in Figures \ref{fig:coverage-gaussian} and \ref{fig:coverage} indicate, apart from the variation between different data sets, that even when a heuristic performs well on average there are often nodes for which the candidate parents do not cover a proportionate part of the posterior mass (e.g., Figure~\ref{fig:coverage}(f)). 
%


\section{DAG sampling and causal effect estimation}

Here we describe in detail the algorithms used for estimating causal effect and discovering ancestor relations. We also describe how the data was generated and how the performance of the algorithms was evaluated. We present further results and discuss some implementation practicalities.

\comment{
\subsection{Model parameters}



}

\subsection{Tested methods}


We first describe the hyperparameters and implementation particulars of our novel methods, and then previous methods. We also present further simulation results.

\paragraph{Hyperparameters of Bayesian models}

Unless noted otherwise, we set the hyperparameters of the priors as follows. 
For continuous data we use BGe (i.e., a normal--Wishart prior) with 
$\alpha_{\mu} = 1$, $\alpha_{w} = n+2$, and $T=\frac{1}{2}I_n$ 
as default in Gobnilp~\cite{cussens:2011}.
For discrete data we employ BDeu with equivalent sample size 10. 
As described in Section~2 of the main paper, we set the prior probability of a DAG  proportional to $1 \big/ \prod_{i=1}^n \binom{n-1}{d_i}$, where $d_i$ is the number of parents of node $i$ in the DAG. These choices ensure that Markov equivalent DAGs receive the same score, i.e., posterior probability; while we regard this result as a folklore, we include the following proof for completeness:

\medskip   
\begin{proposition}
The multiset of node indegrees is unique for DAGs in the same equivalence class. 
\end{proposition}
\vspace*{-12pt}
\begin{proof}
An edge $i \rightarrow j $ is called \emph{covered} in a DAG $G$ if $\parm_G(j)=\parm_G(i) \cup \{i\}$~\cite[Def.~2]{DBLP:conf/uai/Chickering95}. Consider reversing $i\rightarrow j$ to $i \leftarrow j$ in $G$ to form $G'$. Because a covered edge cannot  be a part of an unshielded v-structure, we have that $G$ and $G'$ are in the same Markov equivalence class. Furthermore, a covered edge reversal does not change the multiset of node indegrees, since the indegrees of nodes $i$ and $j$ are simply switched.
Now, by Theorem~2 of Chickering~\cite{DBLP:conf/uai/Chickering95}, one can move through all DAGs in a Markov equivalence class by a sequence of covered edge reversals.
Hence, since the multiset of node indegrees remains unaltered in any single covered edge reversal, all members of a Markov equivalence class must have the same multiset of node indegrees.
\end{proof}

\paragraph{Our methods}

\begin{itemize}[leftmargin=64pt]

\item[\gadget{}] For selecting candidate parents, \gadget{} uses \emph{Greedy-lite}. \added{The number of candidate parents $K$ was set as large as possible such that computations other than MCMC iterations took at most a half of the allowed time budget. The running time performance of the different parts of the system was estimated for each input by a short preliminary test run.} The first 50 \% of the iterations were disregarded as burn-in, and thinning was set to obtain $10~000$ DAGs.

\item[\baies{}]  This essentially implements Algorithm~2 of the main paper in R. \baies{} can utilize DAGs sampled by either  \gadget{} or  \bidag{}. The employed normal--Wishart prior is the same as used for sampling DAGs. 
\end{itemize}

\paragraph{Previous methods for averaging over DAGs}

\begin{itemize}[leftmargin=64pt]

\item[\bidag{}] We use the {\tt \small partitionMCMC} function implemented in the BiDAG R-package~\cite{Kuipers2020}. \added{ The algorithm determines its sampling space and the number of candidate parents automatically using a search.} In some of the $107$-variable runs, {\tt \small partitionMCMC} exits and suggests to increase {\tt \small HARDLIMIT} of the allowed number of possible parents ($K$). In these cases we reran {\tt \small partitionMCMC} with an increased {\tt \small HARDLIMIT}.  The first 50~\% of the iterations were disregarded as burn-in, and thinning was set such that we obtain $10~000$ DAGs.

\item[\beandisco{}] We use the authors' implementation available online~\cite{Niinimaki16}. For $20$ and $50$ variables, the maximum number of parents was set to $5$ and $4$, respectively. Note that \beandisco{} employs a so-called order modular graph prior, which results in a posterior that is not score equivalent. The first 50 \% of the iterations were disregarded as burn-in.

\item[\emph{Exact}] This is the ARP algorithm of Pensar et al.~\cite{Pensar2020}, which computes the exact posterior of ancestor relations using dynamic programming and inclusion--exclusion recurrences.
\end{itemize}

\begin{figure}[t!]
\centering
\hspace*{-10pt}
  \begin{subfigure}[t]{.33\textwidth}
\includegraphics[height=100pt,trim={0 0 0 0},clip]{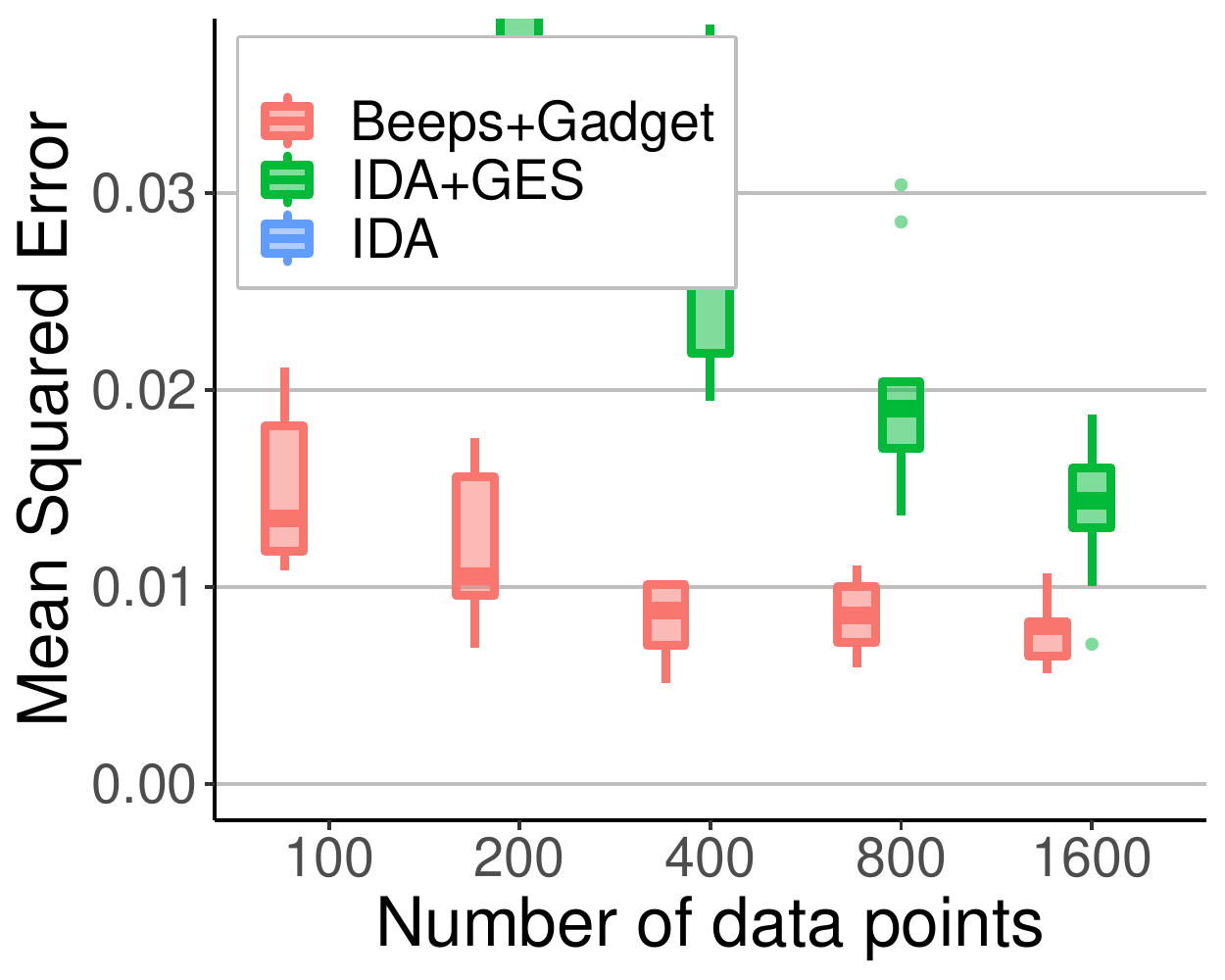} 
    \caption{46-node ECOLI70}\label{fig:ecoli} 
  \end{subfigure}    
\hspace*{-10pt}
  \begin{subfigure}[t]{.33\textwidth}
\includegraphics[height=100pt,trim={0 0 0 0},clip]{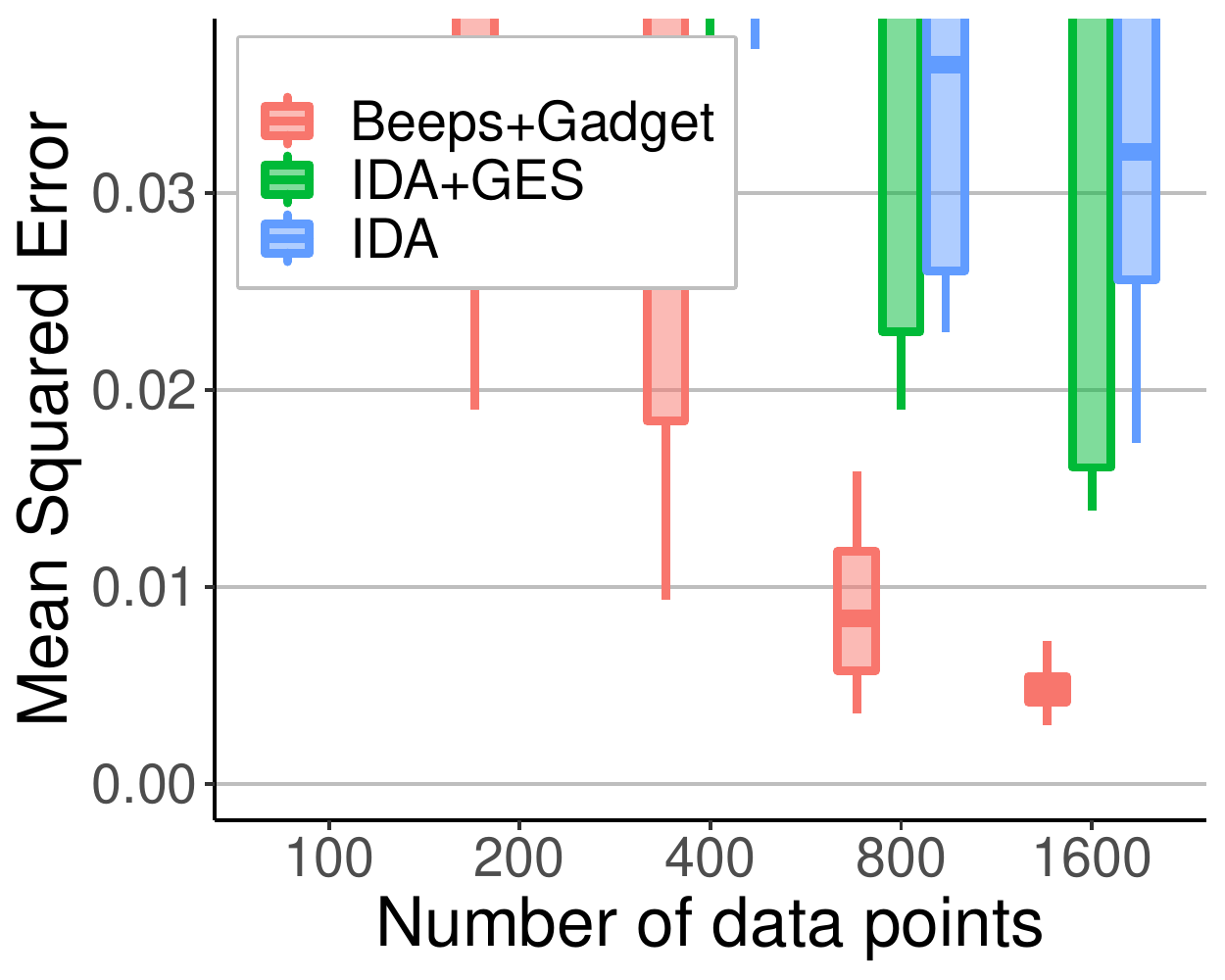} 
    \caption{44-node MAGIC-NIAB}\label{fig:niab} 
  \end{subfigure}      
\hspace*{-10pt}
  \begin{subfigure}[t]{.33\textwidth}
\includegraphics[height=100pt,trim={0 0 0 0},clip]{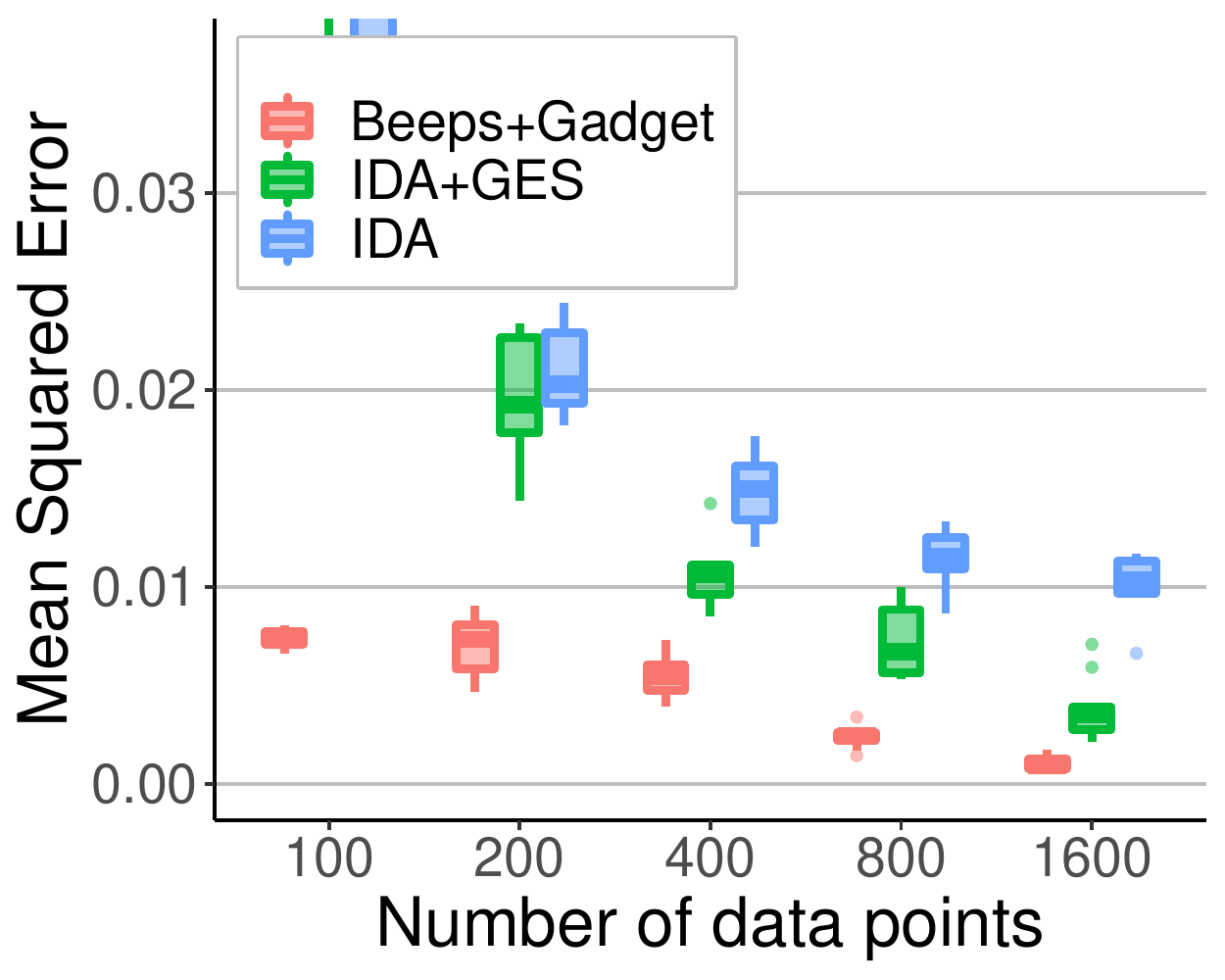} 
    \caption{64-node MAGIC-IRRI}\label{fig:irri} 
  \end{subfigure}
\vspace{12pt}\\
    \begin{subfigure}[t]{.4\textwidth}
\hspace{20pt}\includegraphics[height=100pt,trim={0 0 0 0},clip]{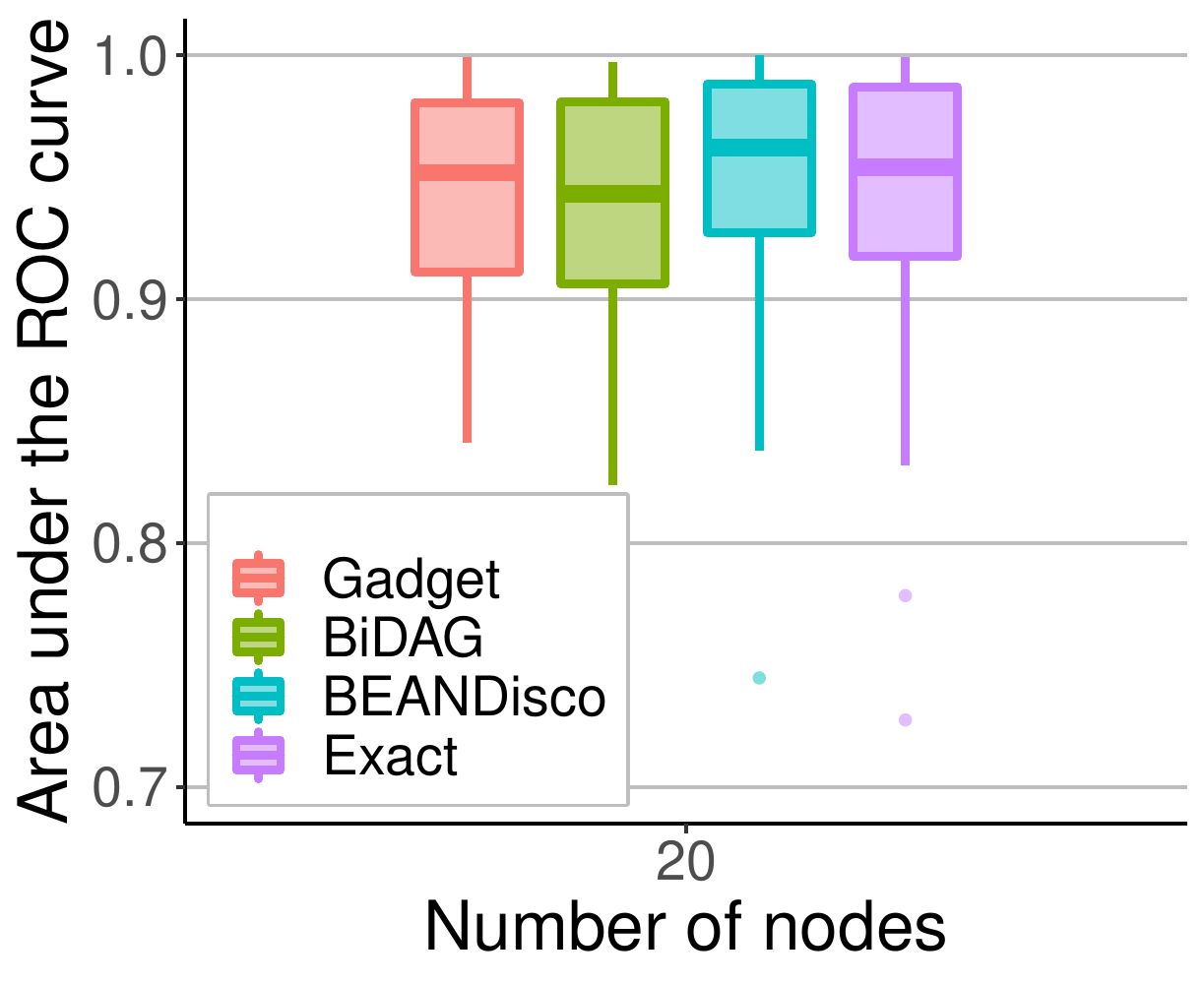} 
    \caption{Ancestor relations for Gaussian data
    }\label{fig:gaussiananc} 
    \end{subfigure}
    \begin{subfigure}[t]{.59\textwidth}
\includegraphics[height=100pt,trim={0 0 4cm 0},clip]{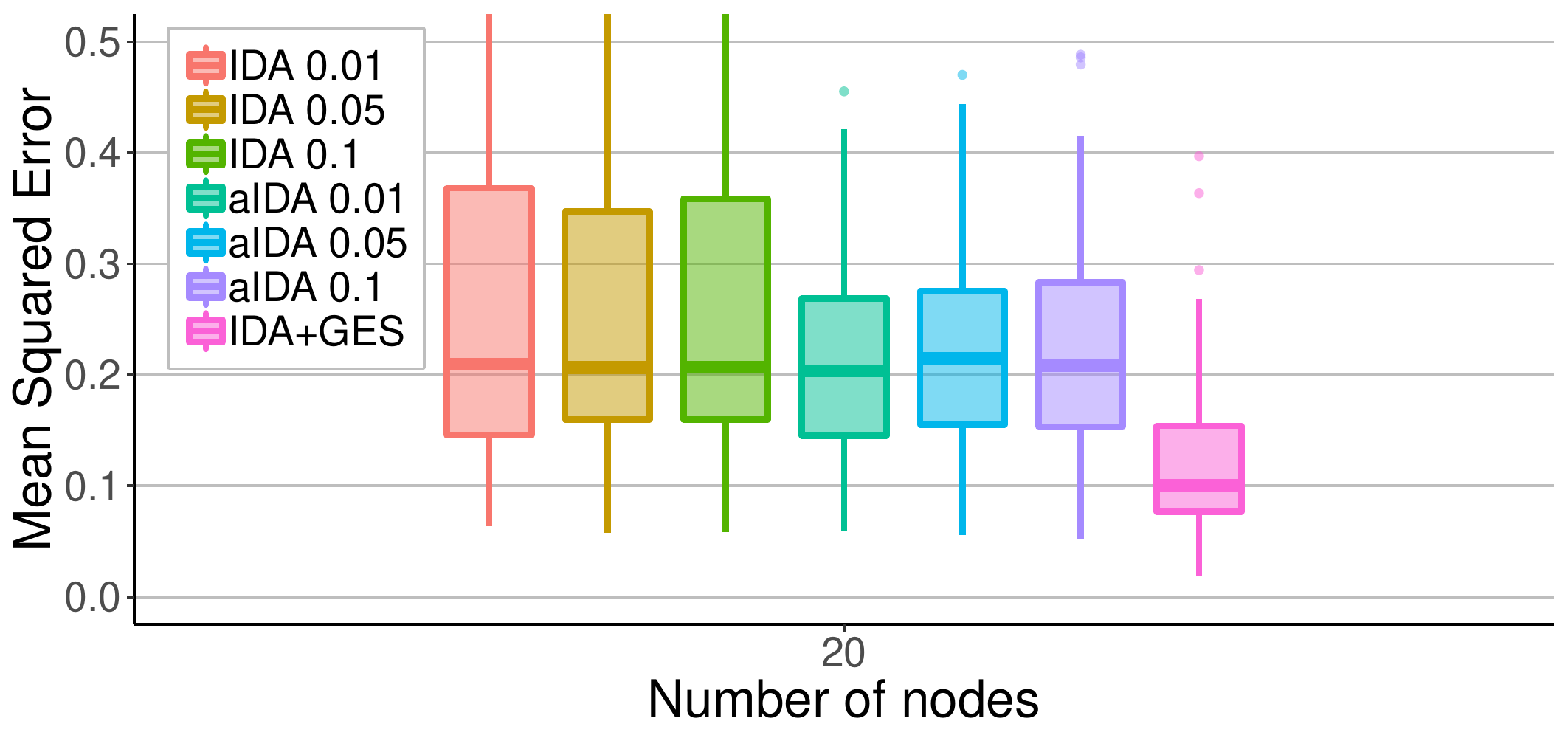} 
    \caption{ The effect of different p-value thresholds on \emph{IDA} and \emph{aIDA}}\label{fig:pval} 
  \end{subfigure}    
\caption{Further performance comparisons. The MCMC methods were run for $1$ hours for $20$-node data and $3$ hours for the benchmarks.} 
\label{fig:mainplots}
\end{figure}

\paragraph{Methods based on IDA}

\begin{itemize}[leftmargin=64pt]

\item[\bida{}] For BIDA we use the available original implementation with the default priors and scores mentioned in the paper~\cite{Pensar2020}. In particular, they employ a fractional marginal likelihood based score.

\item[\emph{IDA}] We use IDA and PC from the pcalg package~\cite{kalisch:2012}. The p-value threshold is set to $0.05$. Figure~\ref{fig:mainplots}(e) shows that the threshold does not have a major effect on the accuracy for the considered datasets.

\item[\emph{IDA+GES}] We use IDA from the pcalg package~\cite{kalisch:2012,IDA}. We employ the GES algorithm in combination with the BIC score, also from pcalg~\cite{kalisch:2012}. IDA has been previously coupled with GES~\cite{Castelletti2020} and other structure learning algorithms~\cite{Pensar2020}.

\item[\emph{aIDA}] We use aIDA with the default settings suggested in the implementation~\cite{Taruttis15}, thus setting p-value threshold of the PC algorithm to $0.1$. Figure~\ref{fig:mainplots}(e) shows that the threshold does not have a major effect on the accuracy for the considered datasets.

\item[\emph{jIDA}]  We test both methods, RRC and MCD, as implemented in the pcalg package~\cite{kalisch:2012,joint}. We employ PC with a p-value threshold 0.05 and GES with BIC for obtaining the Markov equivalence class.
\end{itemize}

Unfortunately, we were not able to get sensible results from the R-code accompanying Castelletti and Consonni~\cite{Castelletti2020} for the data set sizes considered here.

\subsection{Test data}

For the synthetic models, edges were included in the graph randomly such that the average neighbourhood size was 4. The linear Gaussian data were generated as described in the main paper Section~4.2, and standardized to
zero mean and unit variance \cite[Assumption B]{IDA}.
 The true causal 
effects were calculated from the standardized
models. For the discrete case, we considered binary variables and the model parameters were drawn from a Dirichlet with an equivalent sample size (ESS) of $10$. 


\subsection{Empirical results}




\paragraph{Marginal causal effects}

Figures~\ref{fig:mainplots}(a--c)
show the performance of Gadget and the IDA-based methods in estimating causal effects for additional benchmark datasets obtained from the BNLEARN-network repository~\cite{scutari:2010}. \baies+\gadget, with the running time of 3 hours,  is able to provide more accurate estimates, and the accuracy is improved with increasing number of data points.   \emph{IDA+GES} needs twice as many data points  to reach a similar level of accuracy as  \baies+\gadget. Figure~\ref{fig:mainplots}(e) shows that different p-value thresholds do not improve the performance of the methods employing the PC algorithm.

\paragraph{Ancestor relations}

Figure~\ref{fig:mainplots}(d) shows that all MCMC methods are able to closely match the performance of the exact approach in detecting ancestor relations in linear Gaussian data. This is similar behaviour as seen in Figure 3(a) in the main paper for discrete data. Note that ancestor relation posteriors are more accurate in predicting
the presence of  ancestor relations in the true graph, than various applied IDA-based approaches~\cite{Pensar2020}.

\paragraph{Joint causal effects}

In Figure~3(c) in the main paper we evaluated the quality of the estimated causal effects under multiple interventions. We plot the MSE of the estimated causal effects w.r.t. the true ones, where all successive pairs of variables (i.e., $\{x_1,x_2\},\{x_2,x_3\},\ldots, \{x_{n-1},x_n\}$) are intervened on and we consider all causal effects of the intervened variables on the remaining variables.

\begin{figure}[t!]
\centering
    \begin{subfigure}[t]{.4\textwidth}
\hspace{20pt}\includegraphics[height=100pt,trim={0 0 0 0},clip]{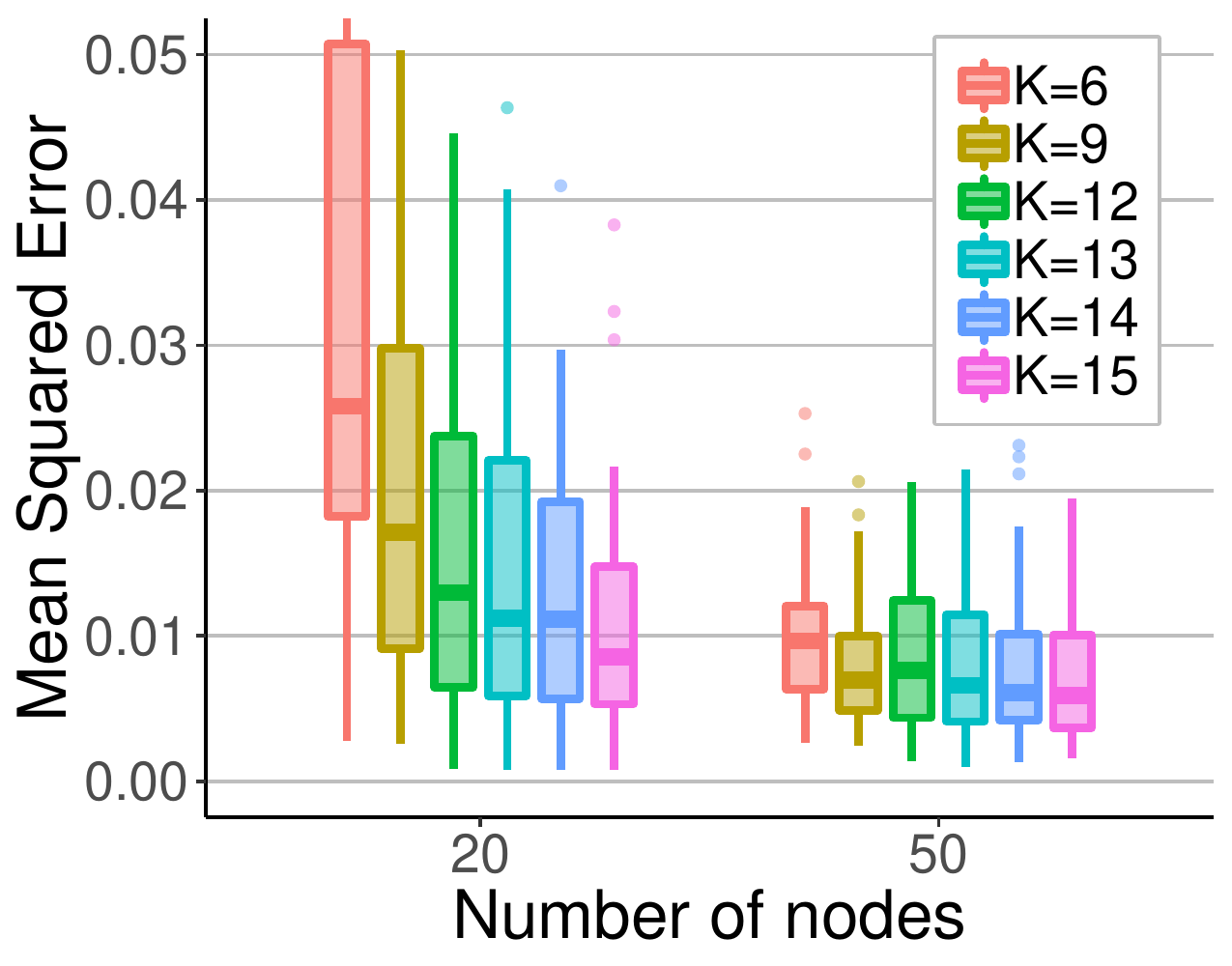} 
    \caption{
   Marginal causal effects for Gaussian data.
    }
    \end{subfigure}
    \begin{subfigure}[t]{.4\textwidth}
\hspace{20pt}\includegraphics[height=100pt,trim={0 0 0cm 0},clip]{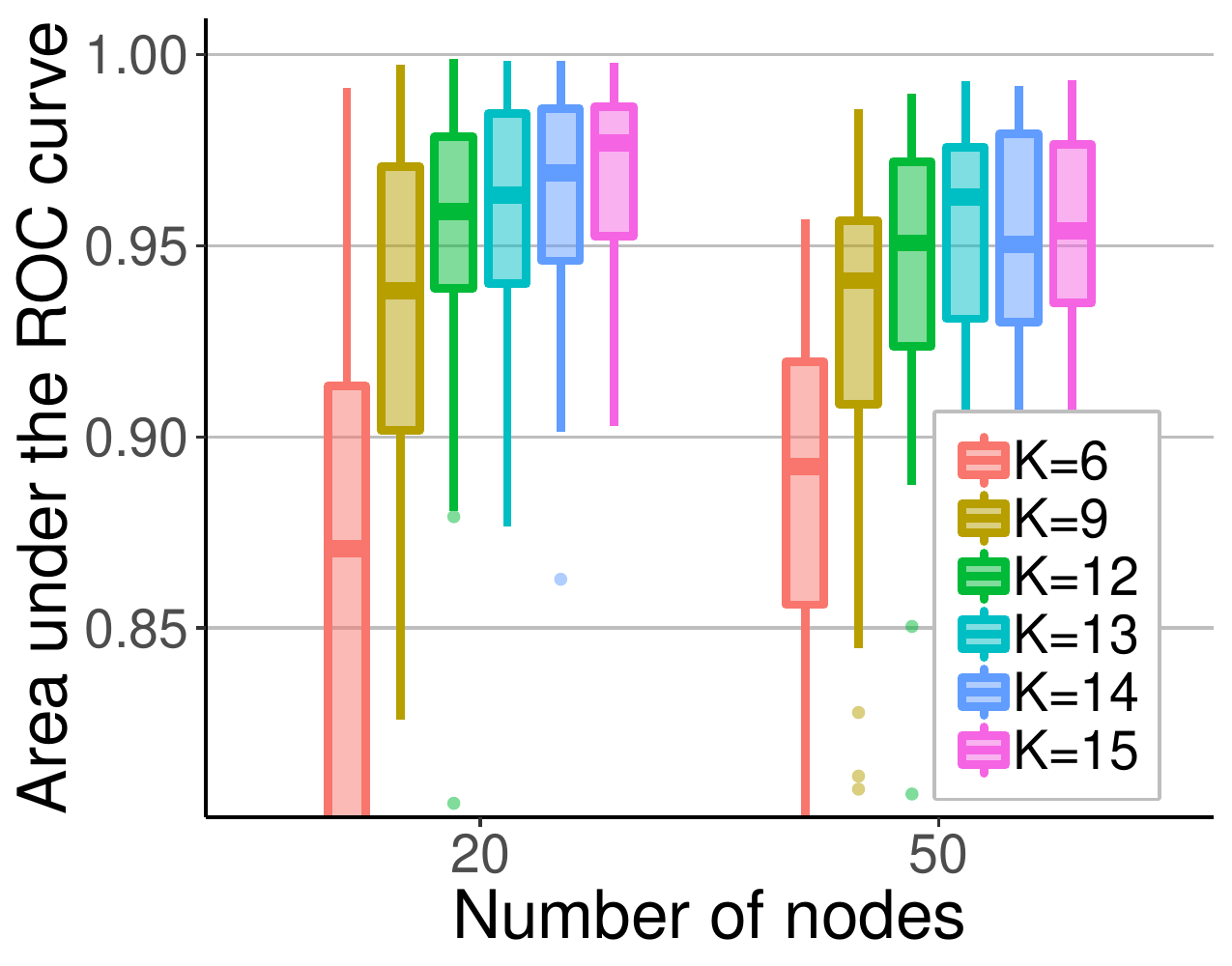} 
    \caption{ Ancestor relations for Gaussian data.}
  \end{subfigure}    
\caption{\added{ The effect of the number of candidate parents ($K$) on estimation performance. The MCMC methods were run for $1$ hours for $20$-node data and $3$ hours for the $50$-node data.}} 
\label{fig:Kplots}
\end{figure}

\added{
\paragraph{The effect of the number of candidate parents ($K$)} Figure~\ref{fig:Kplots} shows the accuracy performance of \baies{}+\gadget{} when using different values for the number of candidate parents $K$. 
These results indicate that larger $K$ values generally produce better accuracy performance. For 50 variable the higher $K$ values mean shorter MCMC chains are possible within the time budget of 3 hours---this results in a slight drop in accuracy for detecting ancestor relations in Figure~\ref{fig:Kplots}(b). }

\begin{figure}[t!]
\begin{center}

\center
\includegraphics[scale=1]{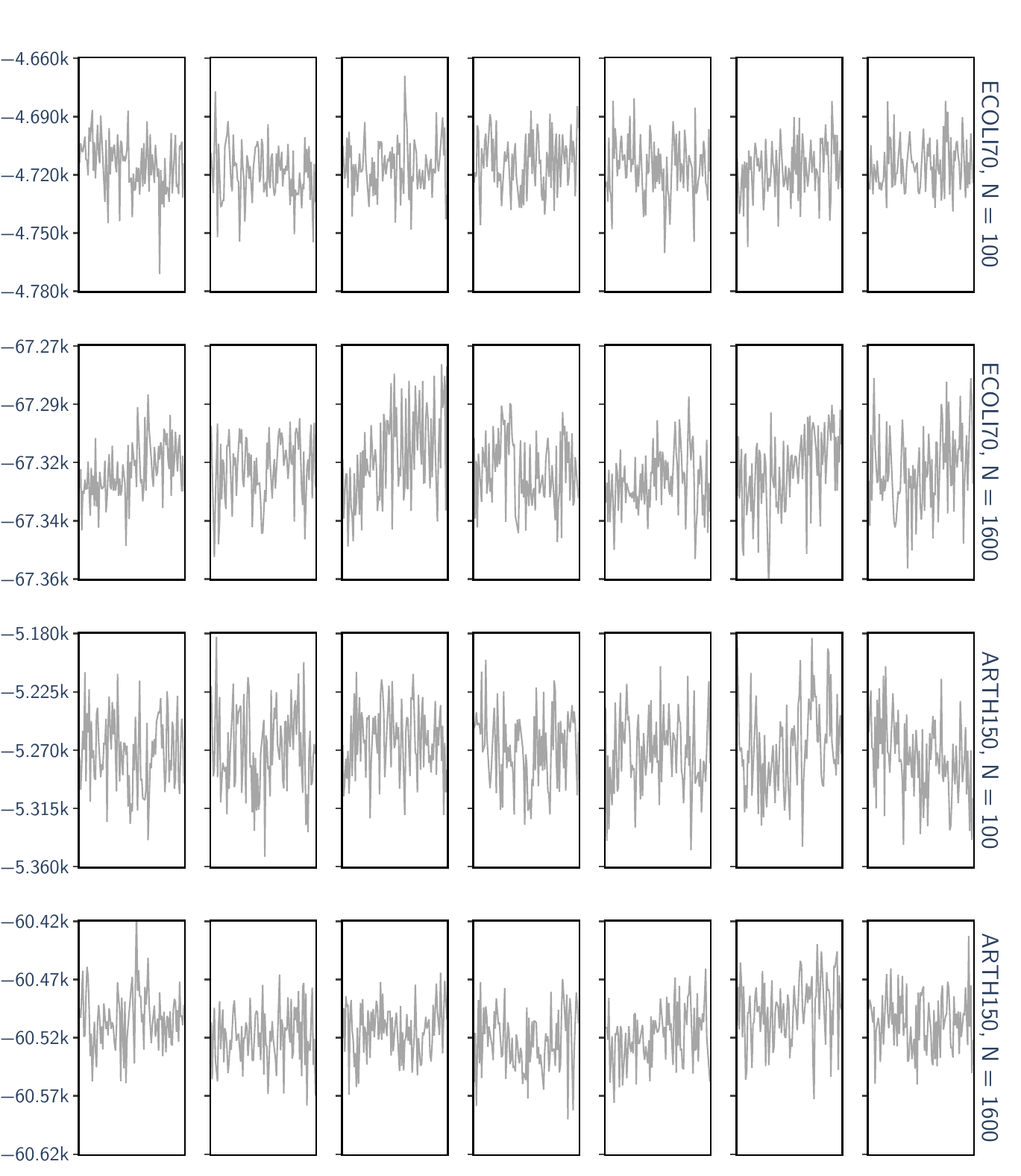}

\end{center}
\caption{Mixing of Gadget on data sampled from two benchmark BNs. $Y$-axis shows the posterior probability of the sampled DAGs (a logarithm of the unnormalized posterior). $X$-axis represents the simulation steps after the burn-in phase, during which $100$ evenly spaced DAGs were sampled. The columns show the results for 7 independent runs. Burn-in was set to $50 \%$ of the chain length, and the running time was set to 3 hours for ECOLI70 and 12 hours for ARTH150.
}
\label{fig:mixing}
\end{figure}

\added{
\paragraph{Mixing of Markov chains}
Figure~\ref{fig:mixing} shows the mixing performance of \gadget{} on datasets of $100$ and $1600$ data points sampled from two benchmark Gaussian BNs from the BNLEARN-network repository~\cite{scutari:2010}. The networks, ECOLI70 and ARTH150, specify a distribution on $46$ and $107$ nodes, respectively. The running times were 3~hours for ECOLI70 (as in Figure~\ref{fig:mainplots}), and 12~hours for ARTH150 (as in Figure~3 in the main paper). The 7 independent runs for each data set reach similar levels of posterior probability, with similar variance, indicating good mixing performance. 
}

\begin{table}[t!]
\centering
\caption{Examples of the share of running time among the components of \gadget{} and \baies{}. Note that the running times spent on different components depend on which $K$ and number of iterations are used. The discrete data runs did not include any computations for causal effects.}\label{tab:times}  
\medskip
\begin{tabular}{rrrr}
\toprule  
Data type                                         & Gaussian & Discrete & Gaussian \\
Number of nodes $n$                               & 20~       & 50~       & 107~      \\
\smallskip
Number of candidate parents $K$                   & 14~       & 14~       & 15~       \\
Search for candidate parents                      & $<$ 1 \%     & $<$ 1 \%     & 1 \%     \\
Precomputing for MCMC                             & 3 \%     & 3 \%     & 9 \%     \\
MCMC iterations                                   & 79 \%    & 86 \%    & 69 \%    \\
Precomputing for DAG sampling                     & 14 \%    & 9 \%     & 17 \%    \\
DAG sampling                                      & 2 \%     & 2 \%     & 4 \%     \\
\smallskip
Computing causal effects                          & 1 \%     & --~        & $<$ 1 \%     \\
Total time                                        & 1 h~      & 3 h~      & 12 h~     \\
\bottomrule
\end{tabular}
\end{table}


\paragraph{Running time performance} Table~\ref{tab:times} reports example running times for the different parts of our methods. Most time is spend in pre-computation, and in sampling root-partitions and DAGs.

\paragraph{Infrastructure} The experiments were run in computer clusters employing Intel Xeon E5-2680 v4 processors.

\comment{
\section{Errors that need to fixed (remove this section before submission)}

Typical minor stylistic errors in writing that need to be corrected:

\begin{enumerate}
\item ``in \cite{gh}.'' $\rightarrow$ ``in Geiger and Heckerman \cite{gh}.''
\item ``(a-c)'' $\rightarrow$ ``(a--c)''
\item ``Figure 1c''  $\rightarrow$ ``Figure~1(c)''
\item ``3h''  $\rightarrow$ ``3~h''
\item ``$20 000$''  $\rightarrow$ ``20~000''
\item In foonotes, always use complete sentences.
\item ``didn't''  $\rightarrow$ ``did not''
\item ``foo -- junk''  $\rightarrow$  ``foo---junk'' (common in Am English)
\end{enumerate}
}

\bibliographystyle{plain}
{\small
\bibliography{paper}
}